\pdfoutput=1
\documentclass[10pt,twocolumn,letterpaper]{article}
\usepackage[T1]{fontenc}
\usepackage{cvpr}
\usepackage{graphicx,amsmath,amssymb,booktabs,multirow,array}
\usepackage{makecell}
\usepackage{tikz}
\usetikzlibrary{arrows.meta,positioning}
\definecolor{cvprblue}{rgb}{0.21,0.49,0.74}
\usepackage[pagebackref,breaklinks,colorlinks,allcolors=cvprblue]{hyperref}

\title{Acting on Recorded Events: Agentic Viewpoint Authoring via Dynamic Target Grounding and Correction}
\author{Longjun Luo$^{1,*}$ \quad 
Weilong Yan$^{2,*}$ \quad 
Hongbin Xu$^{3}$ \quad 
Gang Xu$^{4}$\\
Liangqiong Qu$^{5}$ \quad 
Yongyong Chen$^{6}$ \quad 
Ming Li$^{7,\dagger}$\\[3pt]
{\normalsize $^{1}$Guangdong University of Technology \quad $^{2}$National University of Singapore}\\
{\normalsize $^{3}$ByteDance Inc. \quad $^{4}$Guangming Laboratory \quad $^{5}$University of Hong Kong}\\
{\normalsize $^{6}$Harbin Institute of Technology \quad $^{7}$The Chinese University of Hong Kong Shenzhen}
}
\begin{document}
\AddToHookNext{shipout/foreground}{\put(0,0){\hspace*{22mm}\raisebox{-16mm}{\includegraphics[width=68mm]{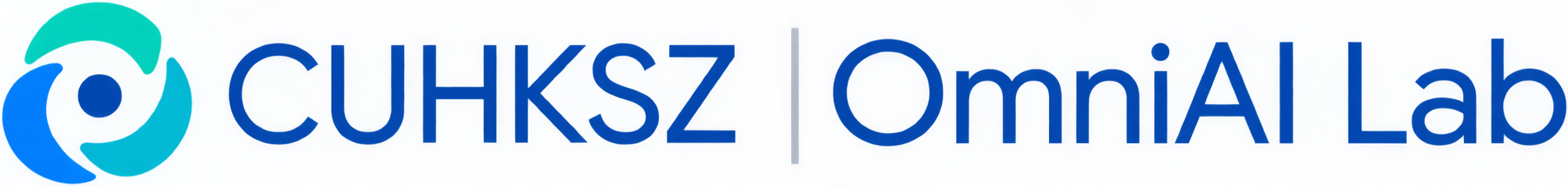}}}}
\twocolumn[{
\renewcommand\twocolumn[1][]{#1}
\maketitle
    \begingroup
\centering
\includegraphics[width=0.95\textwidth]{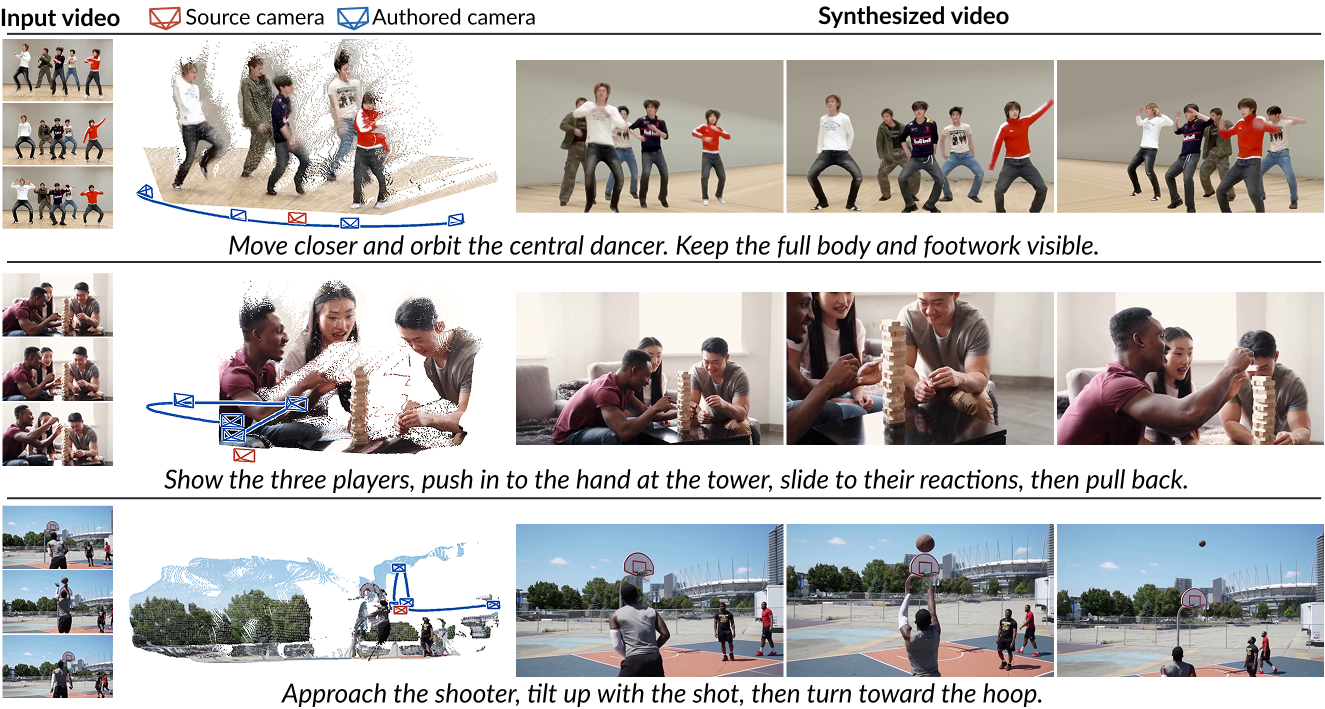}
\vspace{-3mm}
\makeatletter\def\@captype{figure}\makeatother
\caption{\textbf{From filming intent to new observations.} Our agent creates new views of recorded dance, block-game, and basketball scenes in response to filming instructions. Across these scenes, our agent translates filming intent into coordinated changes in subject emphasis, framing, and timing, revealing the actions and interactions the user wants to see.}
\label{fig:teaser}
\par
\endgroup

\vspace{0pt}
}]
\begingroup
\renewcommand{\thefootnote}{\fnsymbol{footnote}}
\footnotetext[1]{Equal Contribution.}
\footnotetext[2]{Corresponding author.}
\endgroup
\flushbottom
\begin{abstract}

4D world models can synthesize new views of the same event along a specified camera trajectory.
Choosing that trajectory from a filming instruction requires understanding what the viewer should see and when it matters. 
We study \textbf{intent-driven viewpoint authoring}, in which users take a monocular recording and a filming instruction as input and produce a camera trajectory with a new video of the same event. 
In our pipeline, an agent grounds each stage of the instruction in dynamic targets tracked through the recording. 
At each key moment, it specifies what the camera should follow and where it should aim, then inspects candidate views to select a composition.
After a compiler connects the selected views, a geometric reviewer projects the same targets through the camera sequence to identify where framing breaks between key moments. 
When the observations support a correction, the system adjusts the camera controls and rechecks the trajectory before synthesis. 
What's more, we introduce 4D-FilmIntent, a benchmark of 160 videos with a corresponding explicit filming-intent task for each scene, with evaluation centered on task fulfillment and synthesized video quality. 
Experiments show that our method fulfills filming intent much better than the evaluated baselines while maintaining competitive video quality.

\end{abstract}

\section{Introduction}

An important goal of 4D world modeling is to recover dynamic scene representations from limited video observations, allowing the same event to be observed from different viewpoints.
Recent reconstruction and video synthesis methods increasingly support this capability along specified camera trajectories~\cite{bai2025recammaster,yu2025trajectorycrafter,lin2026vista4d,meridian}.
Yet generating new views does not determine which views serve a user's intent.
In the block-game example of Figure~\ref{fig:teaser}, the filming intent shifts attention from the group to the hand at the tower and then to the players' reactions. Fulfilling this intent requires selecting the relevant targets and timing each shift to the unfolding interaction.
We study \textbf{intent-driven viewpoint authoring}: translating a monocular video and a filming instruction into new camera controls and a new observation, with the goal of preserving the recorded event while changing its presentation.

Recent camera-planning methods provide several ways to translate language and cinematographic requirements into camera controls. Learned priors capture recurrent patterns of camera motion~\cite{zhang2025gendop}, while LAMP and Auteur expose structured motion and actor-relative framing interfaces~\cite{kizil2026lamp,kizil2026auteur}. ChatCam and CinemaTraj connect instructions to scene-aware camera operations~\cite{liu2024chatcam,li2026cinematraj}. Visual feedback further supports trajectory-generator training in VERTIGO~\cite{li2026vertigo} and viewpoint search in Look-Before-Move~\cite{bian2026look}. Together, these developments connect semantic requests to executable camera behavior and allow candidate observations to guide the \mbox{choice of viewpoint}.

However, reliance on supplied 3D scenes, structured entities, actor motion, or frame-level visual cues leaves a grounding gap when the input is a real monocular recording. This requires estimating scene geometry from monocular video~\cite{Yan_2025_CVPR} and associating partially observed participants and action regions across time. An instruction can also shift between different scales of the same event: the interacting group, a manipulating hand, and the participants' reactions. Keeping a named person visible therefore does not establish that the requested action is readable at the right moment. Errors must be traced to the affected subject, action region, and time interval before the camera controls can be corrected.

This setting calls for a shared reference that connects intent interpretation, camera execution, and visual feedback throughout the recording. Each stage of an instruction specifies what to follow, where to direct attention, and when to change emphasis. The corresponding targets move as the event unfolds, so a view selected at one instant must remain meaningful between key moments. Reviewing the executed sequence against those same targets makes it possible to identify which observation has failed and which control should change.

To address these challenges, we introduce a grounded 4D video agent that connects event understanding, observation planning, and camera correction through dynamic targets in the source video.
The agent first interprets how the event unfolds, translates the user's filming intent into temporally ordered observation requirements, and binds each requirement to participants, local action regions, or interaction groups.
Using the recovered dynamic scene, it proposes and inspects candidate views to assess how the intended targets are framed at the relevant event moments.
The selected views are then compiled into a continuous camera sequence.
Moreover, source observations also guide feedback and correction.
Checks along the compiled sequence localize framing and coverage failures to specific targets and time intervals, while accounting for the reliability of the underlying observations.
This feedback guides bounded adjustments to camera controls, followed by re-execution and inspection.
Finally, the reviewed observation sequence and the source clip guide the video synthesis grounded on the new camera trajectory.
The resulting process connects scene recovery, event interpretation, and feedback-driven camera control, enabling the recorded event to be re-observed according to the user's intended emphasis.

Existing camera-planning benchmarks use executable or reconstructed 3D scenes with structured scene information~\cite{bian2026look,li2026cinematraj}, while video-reshooting evaluations assess synthesis along prescribed camera trajectories~\cite{bai2025recammaster,yu2025trajectorycrafter}. Evaluating camera planning from filming instructions for real monocular recordings remains underexplored. We introduce 4D-FilmIntent to assess this capability through intent fulfillment and synthesized video quality (Sec.~\ref{sec:benchmark}). Experiments show stronger filming-intent fulfillment than the evaluated baselines while maintaining competitive \mbox{video quality (Sec.~\ref{sec:experiments})}.

Our main contributions are threefold:
\begin{itemize}
\item We formulate \textbf{intent-driven viewpoint authoring} over monocular recordings, where event-level filming intentions guide camera planning and new-view synthesis with the goal of preserving the source event.

\item We develop a grounded 4D video agentic pipeline that binds temporally ordered observation requirements to dynamic targets, using these targets as a shared reference for camera execution and evidence-based, target- and frame-specific correction before synthesis.

\item We propose 4D-FilmIntent, a benchmark covering 160 real videos with scene-specific filming instructions to evaluate camera planning for video reshooting, measuring intent fulfillment and synthesized video quality.

\end{itemize}

\section{Related Work}

\subsection{4D Scene Modeling and Video Reshooting}
Spatial representations support object completion~\cite{Yan_2026_CVPR}, local shape editing~\cite{hu2024cns,hu2026cns}, and view synthesis through image-level geometry in MetaView~\cite{cai2026metaview}, camera conditioning in MotionCtrl and CameraCtrl~\cite{wang2024motionctrl,he2025cameractrl,he2025cameractrlii}, and object layouts in CineMaster~\cite{wang2025cinemaster}. Generative Camera Dolly and ReCamMaster retain recorded dynamics under viewpoint changes~\cite{van2024generative,bai2025recammaster}. Geometric references guide video reshooting in TrajectoryCrafter, Vista4D, and Meridian~\cite{yu2025trajectorycrafter,lin2026vista4d,meridian} and world generation in Proxy2World~\cite{Xu2026Proxy2WorldLT}, with related work exploring trajectory, geometry, and camera-grid conditioning~\cite{lan2026holo4d,wizadwongsa2026sierpinskicam,liu2026tars}. Geometric memory supports cross-view consistency~\cite{ren2025gen3c,chen2026anyrecon}, while BulletTime and VerseCrafter expose world time and camera or object motion~\cite{wang2026bullettime,zheng2026versecrafter}. Video generation has also expanded to autoregressive world modeling and editing~\cite{zhao2026minwm,rui2026biwm,zhou2026editstream} and synchronized audio-video synthesis~\cite{xu2026av}. Our agent executes a filming intention through camera-conditioned video reshooting.

\subsection{Camera Trajectory Generation and Control}
Geometric methods such as Toric Space and Camera-on-rails encode composition constraints~\cite{lino2015intuitive,galvane2015camera}; example-driven cinematography and style-aware keyframing learn camera behavior from film~\cite{jiang2020example,jiang2021camera}. Learned trajectory priors use diffusion, text and character motion, autoregression, or direction and speed~\cite{jiang2024cinematographic,courant2024exceptional,zhang2025gendop,zhou2026unveiling}. LAMP and Auteur expose structured motion and actor-relative framing~\cite{kizil2026lamp,kizil2026auteur}, while Pulp Motion couples human and camera motion through on-screen framing~\cite{courant2026pulp}. These representations supply camera controls; our agent grounds their targets and timing in a recorded event.

\subsection{Agentic Viewpoint Planning}
Structured plans organize agentic video editing and generation~\cite{yu2026aurora,lin2023videodirectorgpt,pan2025modular} and personalized short-drama production~\cite{shi2026one}, while Story3D-Agent and FilmAgent coordinate characters and cinematography in virtual environments~\cite{huang2024story3d,xu2025filmagent}. Language interfaces enable person-specific video question answering in PVChat~\cite{pvchat}, temporal reasoning over dynamic point clouds in 4DPC$^2$hat~\cite{zhang20264dpc}, and scene-aware camera operations in ChatCam and CinemaTraj~\cite{liu2024chatcam,li2026cinematraj}. Feedback guides iterative scientific video analysis in SciEducator~\cite{xu2026scieducator}, preference-based trajectory training in VERTIGO~\cite{li2026vertigo}, and viewpoint search in Look-Before-Move~\cite{bian2026look}. ATAS maintains scene and event memories for revisits and recapture~\cite{wang2026any}. Our agent connects these planning and review operations through dynamic targets recovered from a monocular recording, enabling target- and frame-specific camera correction before synthesis.

\section{Intent-Driven Viewpoint Authoring}
\label{sec:method}
\begin{figure*}[t]
\centering
\includegraphics[width=0.94\textwidth]{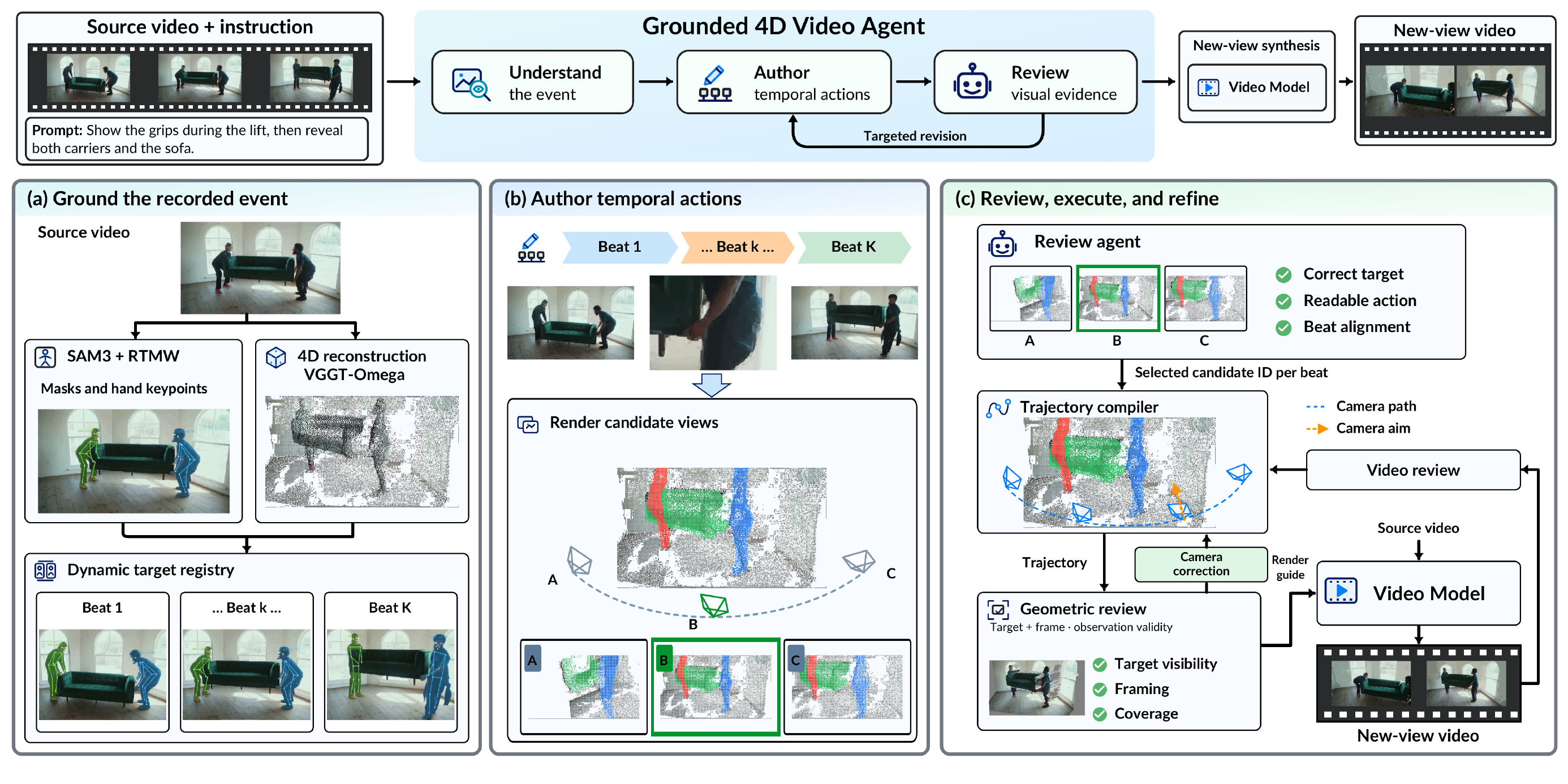}
\caption{\textbf{A 4D video agent for viewpoint authoring.} (a) Source observations ground dynamic targets. (b) Semantic beats specify follow and aim targets; the agent inspects candidate views. (c) The compiler connects selected views, and target- and frame-specific projection review guides bounded corrections before video synthesis. Shared targets connect filming intent to camera execution and correction.}
\label{fig:pipeline}
\end{figure*}

\begin{figure*}[t]
\centering
\includegraphics[width=1.\textwidth]{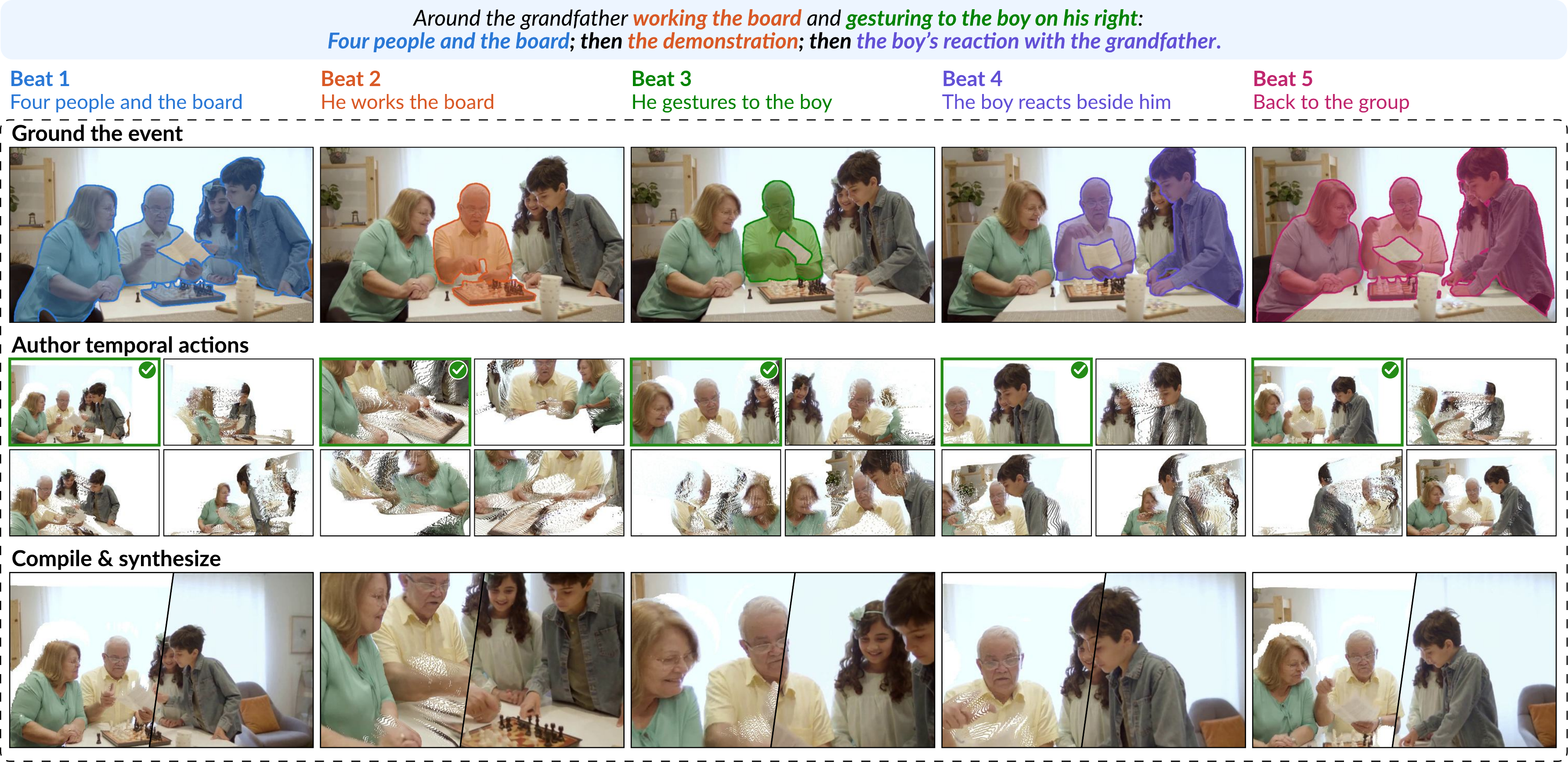}
\caption{\textbf{From filming instructions to event-grounded viewpoints.} Columns show five planned beats. Rows show source frames with beat-colored target masks, four candidate views per beat (green borders and check marks indicate selections), and point-cloud previews (left) paired with synthesized frames (right) at matching source times. The sequence illustrates how our agent selects views for instruction-specified targets and event stages, revealing action details and participant reactions while restoring group context.}
\label{fig:workflow_evidence}
\end{figure*}

\subsection{Overview}
Given a monocular recording $I=\{I_t\}_{t=0}^{T-1}$ and a filming instruction $q$, we seek a camera trajectory $\mathcal{C}=\{(\mathbf{T}_t,\mathbf{K}_t)\}_{t=0}^{T-1}$ and a target video $\hat I$ that shows the same event from the trajectory. 
We write the camera-to-world pose as $\mathbf{T}_t=(\mathbf{R}_t,\mathbf{c}_t)\in SE(3)$ and denote the intrinsics by $\mathbf{K}_t$. 
The instruction determines what the viewer should see as the event unfolds. 
In Fig.~\ref{fig:workflow_evidence}, the instruction calls for the grandfather's demonstration at the board, his gesture toward the boy, and the boy's reaction. 
These moments require different framing even while the participants stay at the same table.

As shown in Fig.~\ref{fig:pipeline}, our agentic pipeline uses a time-indexed scene proxy to make these decisions. 
It grounds the recorded event (Fig.~\ref{fig:pipeline} (a)), binds the intended subject of each shot to a moving target, and inspects candidate views at those moments (Fig.~\ref{fig:pipeline} (b)). 
A compiler connects the selected views, with a geometric reviewer then using the same targets to locate framing errors along the trajectory and guide a correction before synthesis (Fig.~\ref{fig:pipeline} (c)).

\subsection{Grounding Targets in the Recording}
First, grounding a 4D event requires spatial geometry information. 
We use a geometry estimator~\cite{wang2026vggt} to estimate depth $D_t$ and cameras $\mathbf{K}^s_t,\mathbf{R}^s_t,\mathbf{c}^s_t$ of each source frame to construct the point cloud. 
To identify people across frames, we start from identity anchors reviewed on three sampled frames. 
The anchor and head location disambiguate the intended person when several people are present. Points inside the selected mask prompt SAM~3~\cite{carion2026sam} to propagate body and head-region observations through the clip.

We store the targets in a registry $\mathcal{A}$ indexed by persistent ID $i$. Each entry pairs a semantic identity or role $\eta_i$ with its track,
$
\mathcal{A}[i]=\big(\eta_i,(\mathbf{p}_{i,t},r_{i,t},\nu_{i,t})_{t=0}^{T-1}\big),
\label{eq:target_registry}
$
where $\mathbf{p}_{i,t}$ is the center, $r_{i,t}$ is the extent, and $\nu_{i,t}$ is the observation status. 
The ID persists when an observation is missing. 
Robust statistics over the 3D points within a body mask give its center and extent. 
Points within the head region provide a separate aim target. 
Visible wrists from RTMW~\cite{jiang2024rtmw}, lifted with scene depth, locate the hand region for action-detail shots.
We also register objects as targets and use the union of participant masks to define a group. 
When filling and filtering tracks, we retain the original observation coverage and flag suspicious masks and frames that use a fallback head estimate. 
These records let the reviewer identify which projected targets rely on uncertain observations.

\subsection{From Event Beats to Selected Views}
The agent inspects sampled frames across the complete recording and translates $q$ into a storyboard as
\begin{equation}
\mathcal{B}=(b_k)_{k=1}^{K},\qquad b_k=(\tau_k,f_k,a_k,s_k,\ell_k).
\label{eq:grounded_storyboard}
\end{equation}

The key time $\tau_k$ follows the source timeline. Both the follow ID $f_k$ and the aim ID $a_k$ refer to entries in $\mathcal{A}[i]$. 
The shot type $s_k$ and the beat's role $\ell_k$ tell the agent what to look for in a candidate view. 
The follow target anchors camera placement, while the aim target identifies the region the shot should emphasize. 
A demonstration beat can follow the demonstrator while emphasizing the hand region. The following reaction beat can shift the aim to the observer. 
The storyboard also records any focal or motion constraints.

For candidate $j$ at beat $k$, we place the camera around the follow center and orient it toward the aim target:
\begin{equation}
\mathbf{c}_{k,j}=\mathbf{p}_{f_k,\tau_k}
+d_{k,j}\mathbf{B}_{k,j}\,\mathbf{v}(\alpha_{k,j},\beta_{k,j}),
\label{eq:beat_camera_center}
\end{equation}
where $d_{k,j}$ is the camera distance and $\mathbf{v}$ is a unit direction with azimuth $\alpha_{k,j}$ and elevation $\beta_{k,j}$. 
The basis $\mathbf{B}_{k,j}$ expresses this direction in planning coordinates. 
For person targets, the actor-relative branch uses the actor's estimated forward direction, while the source-reference branch uses the recording viewpoint. 
Medium and wide distances offer alternatives for closer framing and shared context. 
The agent compares point-cloud renders at the same event time, inspecting the size and placement of the intended subject, the presence of other required participants or objects, and gaps in the reconstruction. 
It returns one candidate ID per beat, and each ID retrieves the exact camera parameters used to render that preview. 
Fig.~\ref{fig:workflow_evidence} marks the selected preview for each planned beat.

\subsection{Carrying the Selected Views Through Time}
Let $j_k,\ \mathbf{g}_k$ be the selected candidate ID and its look-at point. 
The compiler converts the selected center from Eq.~\eqref{eq:beat_camera_center} and $\mathbf{g}_k$ into two target-relative controls at the key time,
\begin{equation}
\begin{aligned}
\mathbf{o}_k = \mathbf{c}_{k,j_k}-\mathbf{p}_{f_k,\tau_k}, \qquad
\boldsymbol{\delta}_k = \mathbf{g}_k-\mathbf{p}_{a_k,\tau_k}.
\end{aligned}
\label{eq:beat_relative_controls}
\end{equation}

Across beats and transitions, the compiler assembles the follow and aim tracks $\bar{\mathbf{p}}_f(t)$ and $\bar{\mathbf{p}}_a(t)$. 
Adding the interpolated offsets to these moving tracks gives the camera center and look-at point at each source time,
\begin{equation}
\begin{aligned}
\mathbf{c}(t)=\bar{\mathbf{p}}_f(t)+\mathbf{o}(t), \qquad
\mathbf{g}(t)=\bar{\mathbf{p}}_a(t)+\boldsymbol{\delta}(t).
\end{aligned}
\label{eq:compiled_follow_aim}
\end{equation}

For either offset track $\mathbf{y}\in\{\mathbf{o},\boldsymbol{\delta}\}$, let $\Delta_k=\tau_{k+1}-\tau_k$ and $s=(t-\tau_k)/\Delta_k$. 
On this interval we set $\mathbf{y}(t)=H_k(s)$, where $H_k$ is the quintic Hermite polynomial satisfying
\begin{equation}
\begin{aligned}
H_k(e)   = \mathbf{y}_{k+e}, \
H'_k(e)  = \Delta_k\,\dot{\mathbf{y}}_{k+e}, \
H''_k(e) = \Delta_k^2\,\ddot{\mathbf{y}}_{k+e},
\end{aligned}
\label{eq:relative_hermite}
\end{equation}
where $e\in\{0,1\}$.
Primes denote derivatives with respect to $s$ and dots denote derivatives with respect to source time.
Adjacent segments share their knot velocities and accelerations. 
Both derivatives vanish at the first and last knots.
This makes the relative offsets $C^2$ in source time. The world-space path also depends on the moving tracks in Eq.~\eqref{eq:compiled_follow_aim}. 
The direction from $\mathbf{c}(t)$ to $\mathbf{g}(t)$ and world up determine the orientation $\mathbf{R}_t$. 
The compiler plans at a similarity-normalized scale and maps translations back to reconstruction coordinates for projection and rendering.

\subsection{Reviewing and Correcting the Trajectory}
A selected view shows the composition at one beat.
Between beats, the moving subject may approach the image boundary or leave the frame as the camera turns toward the next target. 
At each frame, a deterministic reviewer projects the grounded body, head, and aim observations through the compiled camera.
For a target center $\mathbf{p}_{i,t}$, its camera coordinates $\mathbf{z}_{i,t}$ and image position $\mathbf{u}_{i,t}$ are
\begin{equation}
\begin{aligned}
\mathbf{z}_{i,t}=\mathbf{R}_t^\top(\mathbf{p}_{i,t}-\mathbf{c}_t),\qquad
\mathbf{u}_{i,t}=\pi(\mathbf{K}_t\mathbf{z}_{i,t}),
\end{aligned}
\label{eq:target_reprojection}
\end{equation}
where $\pi(x,y,z)=(x/z,y/z)$ for points in front of the camera. 
The same projection applies to the available body and head observations. 
Their projected extent exposes cropping even when the center lies inside the frame. 
The reviewer also checks aim alignment, proximity to the image boundary, and motion on screen across frames. 
It indexes each issue by target and frame and records the original observation coverage, mask warnings, and head fallbacks. 
A missing observation alone gives no evidence that moving the camera would improve the shot.

Where the observations support a correction, the report triggers at most one bounded update to the camera controls.
A target near the edge can prompt aim recentering. 
Cropping can prompt a larger camera distance or a wider field of view.
Aim misalignment or excessive motion on screen can prompt an adjustment to the aim track or its smoothing. 
Each operation specifies its frame range.
When a task requires a constant field of view, a requested widening sets the widest requested value for the whole clip. 
The compiler regenerates the trajectory, and the reviewer checks it again. 
The execution record retains the original trajectory, the report, and the applied operations.

\subsection{Synthesizing and Reviewing the New View}
Let $\mathcal{C}^{\star}$ denote the trajectory after geometric review and $\mathcal{G}=(\mathcal{G}_t)_{t=0}^{T-1}$ the reconstructed geometry. 
At each source time $t$, we render the point cloud $\mathcal{G}_t$ from the corresponding reviewed camera. 
The resulting sequence $P$ guides composition, while the aligned source clip $I$ supplies appearance and motion cues.
Under synthesis request $\rho$, Meridian~\cite{meridian} produces the new video as
\begin{equation}
P=\mathcal{R}(I,\mathcal{G},\mathcal{C}^{\star}),\qquad \hat I=\mathcal{V}_{\rho}(I,P).
\label{eq:reviewed_synthesis}
\end{equation}
The bottom row of Fig.~\ref{fig:workflow_evidence} pairs the point-cloud preview with the synthesized frame at each beat.

Our post-synthesis review protocol compares the source frame $I_t$, the point-cloud render $P_t$, and the generated frame $\hat I_t$ at the same event time. 
The agent checks whether the output shows the evidence requested for that beat and preserves the participants, objects, and action in the recording. 
It then inspects windows of consecutive frames for errors in action timing, temporal discontinuities, and synthesis artifacts. 
For each issue, the report identifies the affected target and time interval and recommends a change to the camera controls or the synthesis request.
A camera update requires recompiling the trajectory and rendering a new reference sequence. 
A change to $\rho$ alone reuses the current reference. The system automatically resynthesizes the clip with the revised inputs and repeats the review until the output meets the review criteria or the preset retry budget is exhausted.

\section{\texorpdfstring{4D-FilmIntent Benchmark}{4D-FilmIntent Benchmark}}
\label{sec:benchmark}
\begin{figure}[!t]
\centering
\includegraphics[width=\columnwidth]{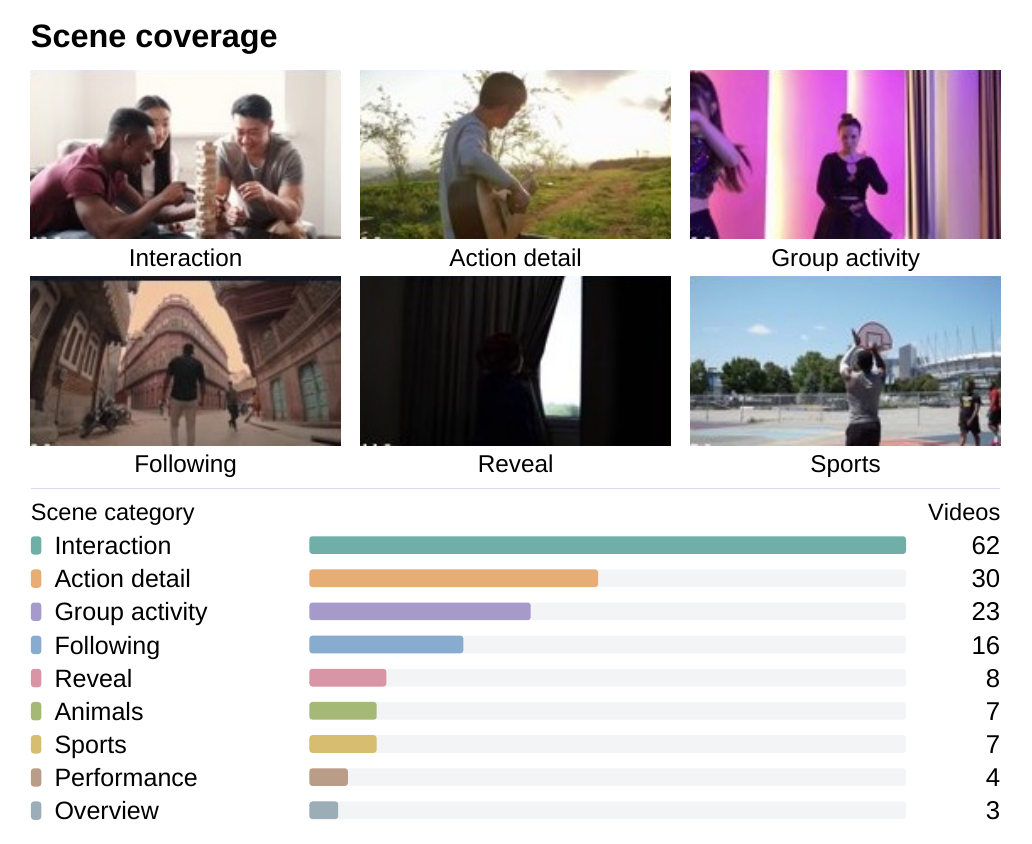}
\caption{\textbf{4D-FilmIntent scene coverage.} Representative clips and the distribution of 160 videos across nine categories.}
\label{fig:benchmark}
\end{figure}
Camera-planning benchmarks in executable or reconstructed 3D scenes provide structured geometry and semantic entities~\cite{bian2026look,li2026cinematraj}, while video-reshooting evaluations assess synthesis along prescribed camera trajectories~\cite{bai2025recammaster,yu2025trajectorycrafter}.
However, supplied scene entities bypass the recovery of participants and action regions from a recording, while prescribed trajectories bypass the decision of where and when to look. Evaluating real-video authoring requires both: recovering the visual evidence of an instruction and choosing how to reveal it over time.
We introduce 4D-FilmIntent to compare how methods plan views from filming instructions for real monocular recordings, where relevant targets and event moments must be grounded in video observations. The benchmark evaluates whether new views present the requested subjects, action details, and interactions at the relevant moments with the required framing, alongside synthesized video quality.

\paragraph{Data selection.}
The videos are sourced from Pexels Videos~\cite{pexels2026videos}, a public stock-video platform. We select 160 clips using source metadata and sampled-frame contact sheets, retaining a source link for each clip. Selection favors recognizable subjects, observable actions, and opportunities to change visual emphasis. The collection combines 100 general scenes with 60 multi-person interactions across nine categories, covering moving subjects, local action details, shared activities, and \mbox{scene context}. Figure~\ref{fig:benchmark} shows representative scenes and the category distribution, with interactions and action details forming the largest groups.

\paragraph{Filming intent.}
We construct each filming-intent task from the source metadata and sampled frames, identifying the subjects, action details, and relationships to emphasize in an ordered sequence. Gemini-3.8-Flash provides sparse source-frame localization of the named participants and objects. Source-frame inspection links these observations to the instruction, forming a task card with stage-specific targets, framing requirements, and interaction groups. For example, an instruction can emphasize the carriers' grips during a lift and then reveal both carriers and the shared object. The task card supplies common observation requirements for evaluating all methods.

    \section{Experiments}
    \label{sec:experiments}
    
    \begin{figure*}[!t]
    \centering
    \includegraphics[width=\textwidth]{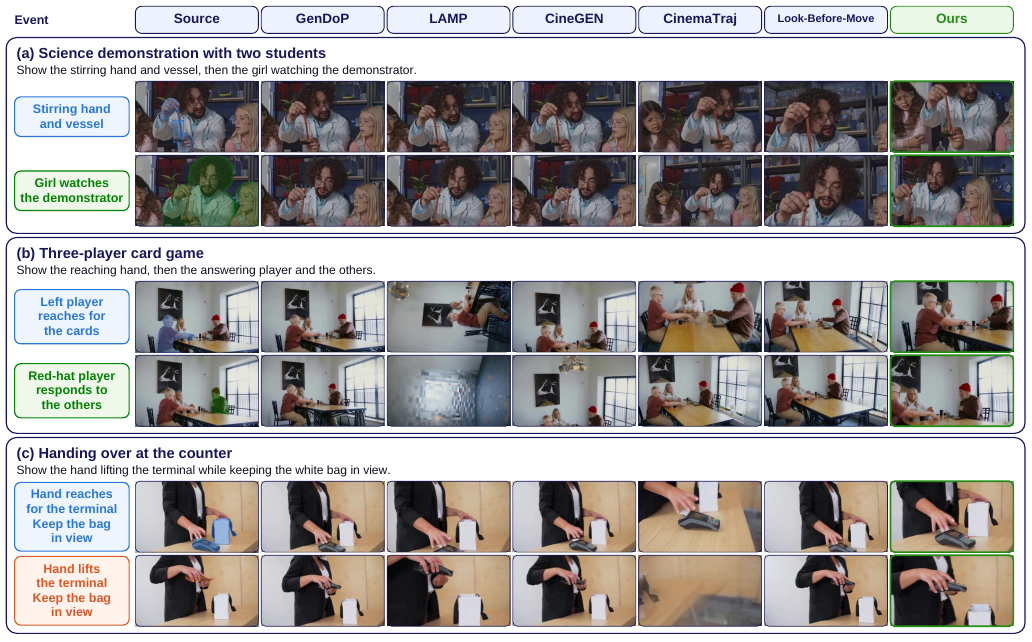}
    \caption{\textbf{Qualitative comparison of instruction-guided filming.} Outputs are compared at matched timestamps. Source masks match the event-label colors, and green borders mark our outputs. Our views emphasize the requested actions and participants while \mbox{preserving interaction context}.}
    \label{fig:pointcloud_comparison}
    \end{figure*}
    We evaluate how well the authored views fulfill filming intent and preserve visual quality, using VLM judgments, geometric measurements, and automatic appearance scores.
    
    \begingroup
    \flushbottom
    \setlength{\parskip}{0pt plus 2pt}
    \subsection{\texorpdfstring{Experimental Setup}{Experimental Setup}}
    \label{sec:metrics}
    \paragraph{Baselines and Implementation.} We compare with GenDoP~\cite{zhang2025gendop}, LAMP~\cite{kizil2026lamp}, CineGEN~\cite{zhou2026unveiling}, CinemaTraj~\cite{li2026cinematraj}, Look-Before-Move~\cite{bian2026look}, CCD~\cite{jiang2024cinematographic}, and ET~\cite{courant2024exceptional}. Each method receives the filming task, and its trajectory is expressed in the reconstructed scene geometry and rendered with Meridian~\cite{meridian}. Our agent uses GPT-5.5~\cite{openai2026gpt55} for temporal authoring and candidate inspection, with SAM~3 masks, RTMW observations, and dynamic target tracks providing scene references. Selected views are compiled into a trajectory and checked for framing corrections before synthesis. We use NVIDIA H20 GPUs for inference. Our video synthesis takes approximately 7.5 minutes per clip.
    
    \paragraph{Evaluation metrics.}
    Filming intent specifies which targets and relationships should be visible and how they should be framed. GPT-5.5~\cite{openai2026gpt55} evaluates anonymized outputs against requirements grounded in the source video. Target coverage measures the fraction of required targets that are readable within each instruction stage. Framing alignment scores the requested composition as absent, partial, or fulfilled. Relation co-framing measures the fraction of sampled frames in which required participants and their relationship are jointly readable. Framing retention measures the fraction satisfying all applicable composition constraints. Visual quality scores each sampled frame for severe corruption, noticeable artifacts, or no damaging artifacts. Scores are normalized to 0--100 and averaged equally across videos; Table~\ref{tab:vlm_judge_results} summarizes the VLM scores.
    
    Non-VLM evaluation uses common source-grounded targets projected through the synthesized camera sequence. CFR measures satisfaction of target retention, scale, and placement constraints; Focus measures central emphasis on the designated target; IC measures joint visibility of required interaction entities; ORDER measures composition fulfillment within successive stage windows. CFR, Focus, IC, and ORDER are reported as fractions. Sharpness consistency measures agreement with source sharpness. Let $r$ denote the reported mean output-to-source ratio of mean Laplacian variance across videos. We define $S_{\mathrm{sharp}}=\min(r,1/r)$ for $r>0$, and $S_{\mathrm{sharp}}=0$ when $r=0$. This transformation is applied to the aggregate ratio; the score reaches one when source and output sharpness agree, and higher is better. VBench Imaging Quality (IQ) and Aesthetic Quality (AQ) measure perceptual quality and visual appeal with the official predictors and preprocessing. Frame scores are averaged within each video and video scores are averaged equally over each method's available outputs; normalized IQ and AQ scores are multiplied by 100 (Table~\ref{tab:pilot}). Higher scores are better for all reported metrics; tied values share the same ranking mark.
    
    \subsection{\texorpdfstring{Quantitative Results}{Quantitative Results}}
    Table~\ref{tab:vlm_judge_results} shows the clearest separation in framing alignment: our agent reaches 82.9, exceeding the strongest baseline by 14.0 points. The qualitative examples in Figure~\ref{fig:pointcloud_comparison} locate this difference in the presentation of task-relevant actions. In the science demonstration, the instruction requires attention to move from the stirring hand and vessel to the observing student. Baseline outputs that retain broad source framing keep the scene recognizable but give limited emphasis to these successive targets; other outputs lose the requested participants during reframing. Our agent binds each instruction stage to a dynamic target and inspects its candidate views before compiling the sequence. Execution review then checks those same targets between key moments. This connection between the instruction, observed action region, and executed view explains the stronger target coverage, relation co-framing, and framing retention.
    \par\endgroup
    
    \begin{table}[t]
    \centering
    \caption{VLM-as-Judge comparison of instruction fulfillment and visual quality on 4D-FilmIntent. Bold and underlined values indicate the highest and second-highest scores.}
    \label{tab:vlm_judge_results}
    \footnotesize
    \setlength{\tabcolsep}{2pt}
    \renewcommand{\arraystretch}{1.1}
    \resizebox{\columnwidth}{!}{%
\begin{tabular}{lccccc}
    \toprule
    \makecell[l]{Method} & \makecell[c]{Target\\coverage $\uparrow$} & \makecell[c]{Framing\\alignment $\uparrow$} & \makecell[c]{Relation\\co-framing $\uparrow$} & \makecell[c]{Framing\\retention $\uparrow$} & \makecell[c]{Visual\\quality $\uparrow$} \\
    \midrule
GenDoP~\cite{zhang2025gendop} & 84.9 & 60.8 & 72.6 & \underline{71.5} & \underline{95.7} \\
LAMP~\cite{kizil2026lamp} & 66.2 & 42.4 & 50.9 & 44.9 & 76.1 \\
CineGEN~\cite{zhou2026unveiling} & \underline{85.4} & 63.7 & \underline{75.7} & 67.1 & 94.6 \\
CinemaTraj~\cite{li2026cinematraj} & 84.1 & \underline{68.9} & 70.4 & 60.4 & 85.5 \\
Look-Before-Move~\cite{bian2026look} & 82.4 & 64.7 & 71.0 & 60.2 & 93.2 \\
CCD~\cite{jiang2024cinematographic} & 33.9 & 20.8 & 17.5 & 16.0 & 55.8 \\
ET~\cite{courant2024exceptional} & 44.1 & 33.6 & 41.5 & 40.1 & 70.3 \\
    \midrule
Ours & \textbf{94.9} & \textbf{82.9} & \textbf{82.4} & \textbf{73.4} & \textbf{97.2} \\
    \bottomrule
    \end{tabular}%
}
    \end{table}
    
    The geometric results in Table~\ref{tab:pilot} reinforce this pattern. Our agent obtains the highest CFR, Focus, and ORDER, with Focus reaching 0.622 versus 0.446 for the strongest baseline. In the card-game example, the required emphasis changes from the reaching player to the responding player, so retaining the group alone leaves part of the instruction unexpressed. Stage-specific follow and aim targets make this change explicit in the camera controls. The grounding ablation in Table~\ref{tab:ablation_geometry} further links dynamic targets to improvements in these three measures. GenDoP and Look-Before-Move achieve higher IC, reflecting stronger joint visibility of the required entities; local action emphasis and simultaneous group visibility impose different framing demands. Our agent also leads the appearance measures. Before synthesis, review identifies target loss and framing errors in the geometric references passed to the shared Meridian backend, connecting the authored composition to \mbox{the synthesized observation}.
    
    \begin{table}[t]
    \centering\footnotesize
    \caption{Geometric composition and synthesized appearance on 4D-FilmIntent. CFR, Focus, IC, and ORDER measure task composition; sharpness consistency, IQ, and AQ measure appearance. }
    \vspace{-2mm}
    \label{tab:pilot}
    \setlength{\tabcolsep}{2pt}
    \renewcommand{\arraystretch}{1.1}
    \resizebox{\columnwidth}{!}{%
\begin{tabular}{lrrrrrrr}
    \toprule
    Method & CFR $\uparrow$ & Focus $\uparrow$ & IC $\uparrow$ & ORDER $\uparrow$ & $S_{\mathrm{sharp}}\uparrow$ & IQ $\uparrow$ & AQ $\uparrow$ \\
    \midrule
GenDoP~\cite{zhang2025gendop} & 0.688 & \underline{0.446} & \textbf{0.803} & 0.646 & \underline{0.731} & \underline{60.06} & \underline{53.15} \\
LAMP~\cite{kizil2026lamp} & 0.393 & 0.224 & 0.348 & 0.415 & 0.610 & 53.10 & 48.15 \\
CineGEN~\cite{zhou2026unveiling} & 0.328 & 0.152 & 0.352 & 0.364 & 0.667 & 57.11 & 51.00 \\
CinemaTraj~\cite{li2026cinematraj} & 0.583 & 0.444 & 0.462 & \underline{0.692} & 0.535 & 53.66 & 49.32 \\
Look-Before-Move~\cite{bian2026look} & \underline{0.693} & 0.443 & \underline{0.726} & 0.596 & 0.642 & 59.35 & 52.46 \\
CCD~\cite{jiang2024cinematographic} & 0.054 & 0.027 & 0.046 & 0.100 & 0.201 & 39.82 & 42.76 \\
ET~\cite{courant2024exceptional} & 0.004 & 0.000 & 0.000 & 0.022 & 0.479 & 46.93 & 46.60 \\
    \midrule
Ours & \textbf{0.713} & \textbf{0.622} & \textbf{0.672} & \textbf{0.731} & \textbf{0.765} & \textbf{63.51} & \textbf{54.38} \\
    \bottomrule
    \end{tabular}%
    \vspace{-3mm}
}
    \end{table}
    
    \subsection{\texorpdfstring{Qualitative Results}{Qualitative Results}}
    Figure~\ref{fig:pointcloud_comparison} illustrates how the authored views follow the requested emphasis. In the science demonstration, attention moves from the stirring hand and vessel to the observing student. In the card game, it shifts from the reaching player to the responding player while retaining the surrounding participants. In the counter scene, the view follows the hand lifting the terminal and keeps the white bag visible. Several baseline outputs retain broad source framing or lose the requested subjects. Our outputs combine readable action details with the context needed to understand the interaction.

    \subsection{\texorpdfstring{Ablation Study}{Ablation Study}}
    \label{sec:ablations}
We study dynamic grounding and feedback revision in recorded events, comparing the full agent with variants that remove either component or both. All configurations independently replan with GPT-5.5 under shared task inputs and synthesis settings. Removing dynamic grounding replaces the semantic dynamic target registry with source-only direct camera poses at a common geometric scale. Removing feedback retains the first independently authored plan. With feedback enabled, the agent inspects the first rendered result and revises the plan once; the full configuration additionally applies native bounded correction.

Tables~\ref{tab:ablation_measured} and~\ref{tab:ablation_geometry} use the VLM, geometric, and appearance metrics from the main comparison. All variants share source-grounded task cards and evaluation windows. GPT-5.5 performs temporal authoring, visual inspection, and evaluation. Video-level scores receive equal weight, with geometric metrics evaluated on \mbox{their common valid inputs}.

\begin{table}[t]
\centering\footnotesize
\caption{VLM-as-Judge ablation of dynamic grounding and feedback revision on 4D-FilmIntent. }
\vspace{-3mm}
\label{tab:ablation_measured}
\setlength{\tabcolsep}{2pt}
\renewcommand{\arraystretch}{1.1}
\resizebox{\columnwidth}{!}{%
\begin{tabular}{@{}lrrrrr@{}}
\toprule
\raisebox{0.5\normalbaselineskip}[0pt][0pt]{Variant} & \shortstack{Target\\coverage $\uparrow$} & \shortstack{Framing\\alignment $\uparrow$} & \shortstack{Relation\\co-framing $\uparrow$} & \shortstack{Framing\\retention $\uparrow$} & \shortstack{Visual\\quality $\uparrow$}\\
\midrule
Full & \textbf{94.9} & \textbf{82.9} & \textbf{82.4} & \underline{73.4} & \textbf{97.2} \\
w/o dynamic grounding & \underline{94.2} & 72.2 & \underline{80.0} & \textbf{74.6} & 93.3 \\
w/o feedback revision & 89.4 & 73.1 & 74.6 & 63.7 & 91.2 \\
w/o both & 87.3 & \underline{76.4} & 72.3 & 68.0 & \underline{95.0} \\
\bottomrule
\vspace{-3mm}
\end{tabular}%
}
\end{table}

\begin{table}[t]
\centering\footnotesize
\caption{Geometric and appearance ablation of dynamic grounding and feedback revision on 4D-FilmIntent. }
\label{tab:ablation_geometry}
\vspace{-3mm}
\setlength{\tabcolsep}{2pt}
\renewcommand{\arraystretch}{1.1}
\resizebox{\columnwidth}{!}{%
\begin{tabular}{@{}lrrrrrrr@{}}
\toprule
Variant & CFR $\uparrow$ & Focus $\uparrow$ & IC $\uparrow$ & ORDER $\uparrow$ & $S_{\mathrm{sharp}}\uparrow$ & IQ $\uparrow$ & AQ $\uparrow$\\
\midrule
Full & \textbf{0.713} & \textbf{0.622} & \textbf{0.672} & \underline{0.731} & \textbf{0.765} & \textbf{63.51} & \textbf{54.38} \\
w/o dynamic grounding & \underline{0.647} & 0.452 & \underline{0.641} & 0.641 & \underline{0.751} & \underline{63.20} & 53.14 \\
w/o feedback revision & 0.616 & \underline{0.550} & 0.608 & \textbf{0.756} & 0.641 & 60.58 & \underline{53.45} \\
w/o both & 0.571 & 0.479 & 0.607 & 0.615 & 0.661 & 61.63 & 52.20 \\
\bottomrule
\vspace{-3mm}
\end{tabular}%
}
\end{table}

\paragraph{Dynamic grounding.}
With feedback held fixed, Full scores 0.731 on ORDER versus 0.641 without dynamic grounding. Framing alignment rises from 72.2 to 82.9, CFR from 0.647 to 0.713, and Focus from 0.452 to 0.622.

\paragraph{Feedback revision.}
With dynamic grounding retained, feedback improves framing retention by 9.6 points and visual quality by 6.0 points. Target coverage, relation co-framing, CFR, and IC also increase. Inspecting rendered views exposes target loss and misplaced emphasis, allowing the agent to revise the controls before synthesis. ORDER decreases from 0.756 to 0.731; the observed benefit of feedback is stronger retention and visual quality rather than a uniform improvement in stage-window composition.

\paragraph{Combined effect.}
Removing both components yields the lowest target coverage, relation co-framing, CFR, IC, and ORDER. The full configuration achieves the highest framing alignment, Focus, visual quality, imaging quality, and aesthetic quality, together with the highest sharpness consistency score. This pattern links target-specific authoring and review of executed views to complementary gains in composition and synthesized appearance.

\section{Conclusion}
We introduced a grounded 4D video agent for intent-driven viewpoint authoring from monocular recordings. The agent connects filming instructions to dynamic targets, selects compositions through visual inspection, and reviews the compiled trajectory before synthesis. We also introduced 4D-FilmIntent to evaluate how the resulting views express filming intent and render the recorded event. Experiments show stronger fulfillment of filming intent than the evaluated baselines while maintaining competitive synthesized \mbox{video quality}. This work connects scene understanding to executable and reviewable observations, making the presentation of a recorded event an explicit target for \mbox{4D video agents} grounded in recorded visual evidence.

\clearpage
\appendix
\raggedbottom
\section*{Supplementary Material}
\section{Dataset Construction}
\label{sec:benchmark_details}

\subsection{Source Discovery and Screening}
We collect publicly available videos from Pexels Videos~\cite{pexels2026videos} in two source groups. The general group covers moving subjects, action details, changes in scene context, and sustained activities. The multi-person group emphasizes events in which two or more identifiable participants act together or respond to one another. We retain each source identifier and URL with the original video and its screening material.

Source metadata identifies the depicted activity, while five-frame contact sheets provide an initial view of its progression. We screen for recognizable people or objects, a readable action or interaction, and room for a meaningful change in framing. For multi-person scenes, simple co-presence is insufficient: the participants must have a visible relation, such as manipulating a shared object, handing something over, taking turns, or reacting to one another. This screening yields 100 clips from the general group and 60 from the multi-person group. The nine semantic categories defined in the main paper describe the selected scene content independently of the two source groups.

\subsection{Clip Preparation}
For each selected video, we identify a continuous source interval that contains the action needed by its filming task. We sample 124 frames at 24 fps from that interval, preserving source-time order. The preprocessing record stores the original source hash, interval start, source-frame indices, spatial crop, and output resolution. Geometry estimation uses a resized and padded version of these frames; the video supplied to synthesis uses the corresponding crop and camera calibration.

\subsection{Explicit Temporal Filming Tasks}
Each of the 160 videos has one explicit filming instruction formed from the recorded action and the visual evidence available in the source frames. Each instruction describes an ordered progression of three observation stages: establish the relevant subject or group, emphasize a consequential action or detail, and recover the participants, shared object, or scene context when called for. Some tasks are anchored to a visible event; others organize a continuous activity through successive changes in visual emphasis. For example, in a sofa-lift scene, the instruction asks for the carriers' grips during the lift and then a view of both carriers and the sofa. Each task specifies what should be visible and when, while permitting different camera trajectories that satisfy the instruction.

\subsection{Source Records and Evaluation Cards}
\label{sec:annotation_design}
Each video is paired with a \emph{source record} and an \emph{evaluation card}. The source record links the video identifier, frame-time mapping, named participants and objects, sparse localizations, and relevant action intervals. For a stage requiring a local action, it points to the evidence region, such as a hand or shared object; for a group stage, it identifies the participants whose relation must remain readable. Visibility and localization evidence are retained with their source-frame provenance.

The evaluation card translates the filming instruction into stage-specific observable criteria: required targets, applicable stage windows, composition constraints, and relations among participants. A criterion can require a detail at an event moment, simultaneous visibility of interacting people, or a return to context later in the sequence. During evaluation, the judge inspects time-aligned source evidence and anonymized synthesized views, recording whether each criterion is met at its intended moment.

\section{Evaluation Protocol}
\label{sec:supp_metrics}

\paragraph{Units and aggregation.}
Let $i$ denote a source video, $k$ an instruction stage, and $t$ a frame. All compared methods operate on the same processed $124$-frame clips, frame-time mappings, evaluation cards, sampled frame indices, and Meridian synthesis pipeline. Each evaluation card fixes the required targets, relations, composition constraints, and stage windows $W_{ik}$, which correspond to frames $0$--$40$, $41$--$82$, and $83$--$123$. A source-verifiable stage is one whose requested evidence is observable in the recording. Scores are computed per video and then averaged over videos with a valid value for that measure. Requirements that do not apply to an instruction, and targets without a valid source observation, have no denominator contribution. All metrics below are higher-is-better except the raw sharpness ratio, whose target is one.

\paragraph{VLM assessment inputs.}
GPT-5.5~\cite{openai2026gpt55} receives the filming instruction, the fixed evaluation card, and anonymized source and synthesized frames at matched time indices. One structured judgment records target visibility and framing for each stage; a second records relations, composition constraints, and image defects. The card defines which targets and relations are applicable. The judge returns discrete judgments, and the formulas below convert them to video-level scores on $[0,100]$.

\paragraph{Target coverage (TC) and framing alignment (FA).}
Let $K_i$ be the source-verifiable stages and $T_{ik}$ the required targets in stage $k$. The binary indicator $y_{ikj}$ equals $1$ when target $j$ is present and clearly recognizable (rather than marginally visible) in the synthesized frames for stage $k$. TC averages target coverage within each stage, then gives every stage equal weight:
\begin{equation}
\mathrm{TC}_i=\frac{100}{|K_i|}\sum_{k\in K_i}\frac{1}{|T_{ik}|}\sum_{j\in T_{ik}}y_{ikj}.
\end{equation}
The judge assigns framing alignment $f_{ik}\in\{0,1,2\}$: $0$ when the requested view is absent, $1$ when its target is present but the requested composition is incomplete, and $2$ when target and composition both satisfy the stage. The video score is
\begin{equation}
\mathrm{FA}_i=\frac{100}{|K_i|}\sum_{k\in K_i}\frac{f_{ik}}{2}.
\end{equation}

\paragraph{Relation co-framing (RC).}
A relation passes when its participants and the interaction between them are jointly readable in the same output frame. Let $R_{ikt}$ be the set of applicable relations at frame $t$, $r_{iktj}\in\{0,1\}$ the binary pass indicator for relation $j\in R_{ikt}$, $F_{ik}$ the sampled frames in stage $k$, and $K_i^{\rm RC}$ the stages containing applicable relation observations. RC first averages within each stage, then across stages:
\begin{equation}
\mathrm{RC}_i=\frac{100}{|K_i^{\rm RC}|}\sum_{k\in K_i^{\rm RC}}\frac{\sum_{t\in F_{ik}}\sum_{j\in R_{ikt}}r_{iktj}}{\sum_{t\in F_{ik}}|R_{ikt}|}.
\end{equation}

\paragraph{Framing retention (FR).}
The evaluation card specifies composition constraints, such as keeping both actors, a shared object, or the surroundings needed to interpret the action. A frame passes FR only when all its applicable constraints hold simultaneously. With binary constraint judgments $b_{itc}$, applicable constraints $C_{it}$, and frames $A_i^{\rm FR}$ that have at least one applicable constraint,
\begin{equation}
\mathrm{FR}_i=\frac{100}{|A_i^{\rm FR}|}\sum_{t\in A_i^{\rm FR}}\prod_{c\in C_{it}}b_{itc}.
\end{equation}

\paragraph{Visual quality (VQ).}
The judge assigns $q_{it}\in\{0,1,2\}$ to each sampled synthesized frame: $0$ for a severe image defect that obscures the event, $1$ for visible artifacts while the content remains readable, and $2$ for a clear frame without a damaging defect. With $F_i$ denoting the sampled output frames,
\begin{equation}
\mathrm{VQ}_i=\frac{100}{|F_i|}\sum_{t\in F_i}\frac{q_{it}}{2}.
\end{equation}

\paragraph{Projection for geometry-based metrics.}
The geometry metrics lift sparse target boxes from the source record using the median valid reconstruction depth inside the source crop and the source camera parameters. The four box corners are interpolated only between adjacent valid observations and projected through the exact camera trajectory supplied to synthesis. For a projected box at frame $t$, $v_t$ is the retained fraction of its area, $a_t$ its clipped area relative to the image, and $c_t\in[0,1]^2$ its projected center. A frame enters a metric only if its source observation and projection are valid.

\paragraph{Camera-framing rate (CFR).}
Omitting video and stage indices $(i,k)$ for brevity, the frame-level pass indicator for a single target is
\begin{equation}
C_t=\mathbf{1}\!\left[v_t\geq0.8,\ a_t\geq a_{\min},\ c_t\in[0.15,0.85]^2\right],
\end{equation}
where $a_{\min}=0.005$ for a face and $0.003$ for another target region. For a group target, the union box passes when its retention is at least $0.65$ and its clipped image area is at least $0.1$. Let $V_i^{\rm CFR}$ contain valid frames within their assigned stage windows. CFR is
\begin{equation}
\mathrm{CFR}_i=\frac{100}{|V_i^{\rm CFR}|}\sum_{(k,t)\in V_i^{\rm CFR}}C_{ikt}.
\end{equation}
The union test measures overall group framing; IC below checks required participants separately.

\paragraph{Focus.}
Focus is evaluated in stages designating one target for emphasis. Let $d_t=\|c_t-(0.5,0.5)\|_2$. Among comparable competing entities with valid projections, retention at least $0.5$, and area at least $0.003$, let $d_t^*$ be the smallest center distance; set $d_t^*=\infty$ if none qualifies. The frame passes when $v_t\geq0.8$, $c_t\in[0.25,0.75]^2$, and $d_t\leq d_t^*$. With $V_i^{\rm Focus}$ denoting the valid frames in focus-required stages,
\begin{equation}
\begin{aligned}
\mathrm{Focus}_i&=\frac{100}{|V_i^{\rm Focus}|}\sum_{(k,t)\in V_i^{\rm Focus}}F_{ikt},\\
F_{ikt}&=\mathbf{1}[v_t\geq0.8,\ c_t\in[0.25,0.75]^2,\ d_t\leq d_t^*].
\end{aligned}
\end{equation}

\paragraph{Interaction coverage (IC).}
Let $J_i$ be the required people and objects for the interaction and $V_i^{\rm IC}$ the frames in its annotated window where every entity has a valid projection. IC checks simultaneous individual visibility:
\begin{equation}
\mathrm{IC}_i=\frac{100}{|V_i^{\rm IC}|}\sum_{t\in V_i^{\rm IC}}\prod_{j\in J_i}\mathbf{1}[v_{jt}\geq0.5,\ a_{jt}\geq0.003].
\end{equation}
The box for a person covers the full body; the box for an object covers the annotated object.

\paragraph{Stage-window fulfillment.}
ORDER measures sustained composition in the prescribed stage windows. Let $F_{ikt}$ be the Focus pass indicator when stage $k$ requires emphasis, and $F_{ikt}=1$ otherwise. A stage passes if there is a run of six consecutive valid frames in $W_{ik}$ for which $C_{ikt}F_{ikt}=1$:
\begin{equation}
o_{ik}=\max_{t:\,[t,t+5]\subseteq W_{ik}}\;\prod_{u=t}^{t+5}C_{iku}F_{iku},
\end{equation}
\begin{equation}
\mathrm{ORDER}_i=\frac{100}{|K_i|}\sum_{k\in K_i}o_{ik}.
\end{equation}
If any stage lacks a valid target observation, ORDER is undefined for that video. The six-frame run prevents a single favorable frame from satisfying a temporal stage.

\paragraph{Synthesized appearance.}
For a video frame $I$, let $L(I)$ be the variance of its grayscale Laplacian after resizing to $448\!\times\!252$. The sharpness ratio is
\begin{equation}
\mathrm{SR}_i=\frac{|F_i|^{-1}\sum_{t\in F_i}L(I^{\rm out}_{it})}{|F_i|^{-1}\sum_{t\in F_i}L(I^{\rm src}_{it})}.
\end{equation}
A ratio near one indicates similar high-frequency energy between the source and synthesized frames. For the main tables, let $r_m$ be the mean video-level SR for method $m$. The reported sharpness-consistency score is $S_{\mathrm{sharp},m}=\min(r_m,1/r_m)$ when $r_m>0$ and zero when $r_m=0$. Imaging Quality (IQ) and Aesthetic Quality (AQ) use the official VBench predictors and preprocessing. IQ measures image-level fidelity and AQ measures learned visual preference. Predictor scores are averaged over decoded frames within a video and then over videos for a method; the normalized values are multiplied by $100$.

\paragraph{Additional diagnostics.}
Content preservation (CP) is the mean cosine similarity of DINOv2 ViT-L/14~\cite{oquab2023dinov2} register-token features from 16 time-matched source and output frames, measuring visual content agreement across the two sequences. Motion adherence (MA) applies to explicit camera-translation requests and measures output-camera displacement relative to the source camera, normalized by the scene's median depth; a translation clause passes when normalized displacement reaches at least $0.03$ depth units within the valid camera range. Tasks without such a clause have no MA score.

\section{Implementation Details}
\label{sec:supp_impl}
The agent first identifies people, action regions, and shared objects in the recorded event. SAM~3~\cite{carion2026sam} masks and RTMW~\cite{jiang2024rtmw} body observations provide image-space evidence, while the 4D reconstruction supplies depth and source-camera geometry. Dynamic target records bind semantic identities to time-varying positions and local regions such as hands. GPT-5.5~\cite{openai2026gpt55} authors the temporal sequence of attention targets and view intentions in the reported experiments. The trajectory compiler resolves those targets into camera controls, renders point-cloud candidate views, and passes the rendered evidence to the review agent. The review step checks target visibility and framing, requests a bounded correction where needed, and sends the resulting camera sequence and source video to Meridian~\cite{meridian} for new-view synthesis.

\paragraph{Baseline input adaptation.}
We run each baseline through its native input interface on the same source videos and filming tasks, supplying the visual inputs and source-derived scene representations accepted by each method. All predicted camera trajectories are transformed into the common reconstruction coordinate system, rendered on the same point cloud, and synthesized with Meridian~\cite{meridian}.

GenDoP~\cite{zhang2025gendop} takes the source visual input together with the task text directly. LAMP~\cite{kizil2026lamp} uses a source-derived scene description to predict object trajectories and the task text to produce camera tags. For CineGEN~\cite{zhou2026unveiling}, we translate the task's ordered visual targets into explicit camera actions---such as establishing the group, moving toward a local action, and pulling back to reveal participant reactions---to match its camera-motion prompt interface. CinemaTraj~\cite{li2026cinematraj} combines the task text with entities localized in the source video via its shot planner and trajectory executor. Look-Before-Move~\cite{bian2026look} receives a source-derived scene summary, character information, and shot constraints, mapping its plan to candidate views in the reconstructed point cloud.

For CCD~\cite{jiang2024cinematographic}, we supply the task text to predict camera-control parameters, recover camera poses following its Unity implementation, and transform them into the reconstruction coordinates. ET~\cite{courant2024exceptional} conditions on the task text and a single character-motion trajectory: we track each participant in the source video with SAM~3~\cite{carion2026sam}, back-project valid torso-mask pixels using reconstructed depth, take a robust median as the 3D body center, and interpolate missing frames, using the framewise mean across participants as the group trajectory.

\onecolumn
\section{Additional Results}
\label{sec:supp_results}

\subsection{Temporal View Comparisons}
\begin{figure}[!htbp]
\centering
\includegraphics[width=\textwidth,height=0.72\textheight,keepaspectratio]{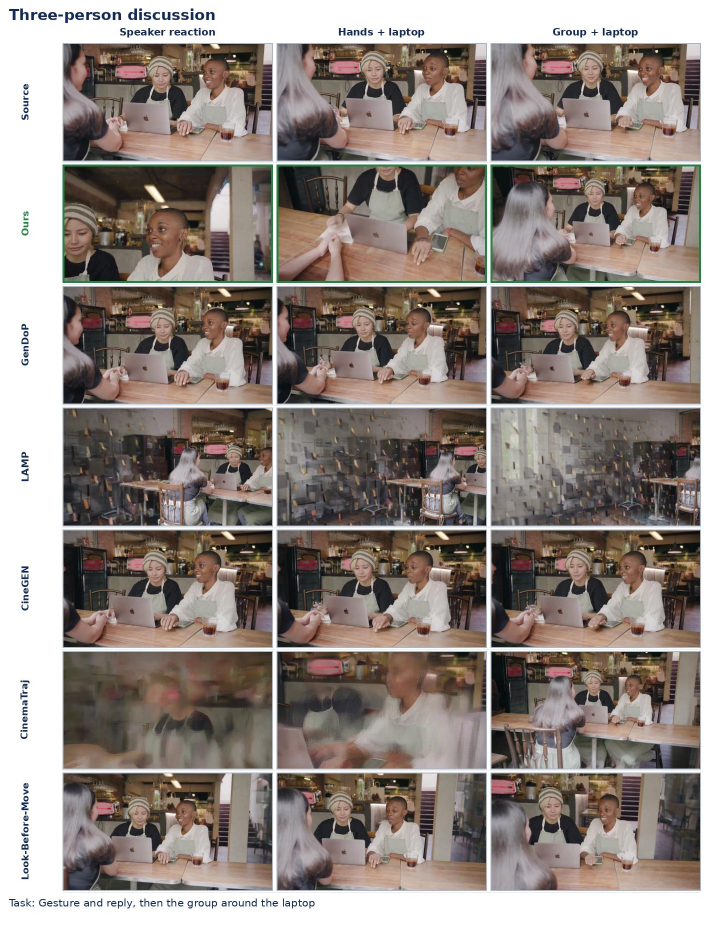}
\vspace{-4pt}
\caption{\textbf{More qualitative comparisons, part 1/10.} We compare our agent with GenDoP, LAMP, CineGEN, CinemaTraj, and Look-Before-Move on a three-person discussion. Columns show matched source times for the speaker's reaction, the hands and shared laptop, and the recovered group. Rows show the source video and each method's synthesized views; green borders mark ours.}
\label{fig:supp_method_4384888}
\end{figure}

\begin{figure}[!htbp]
\centering
\includegraphics[width=\textwidth,height=0.90\textheight,keepaspectratio]{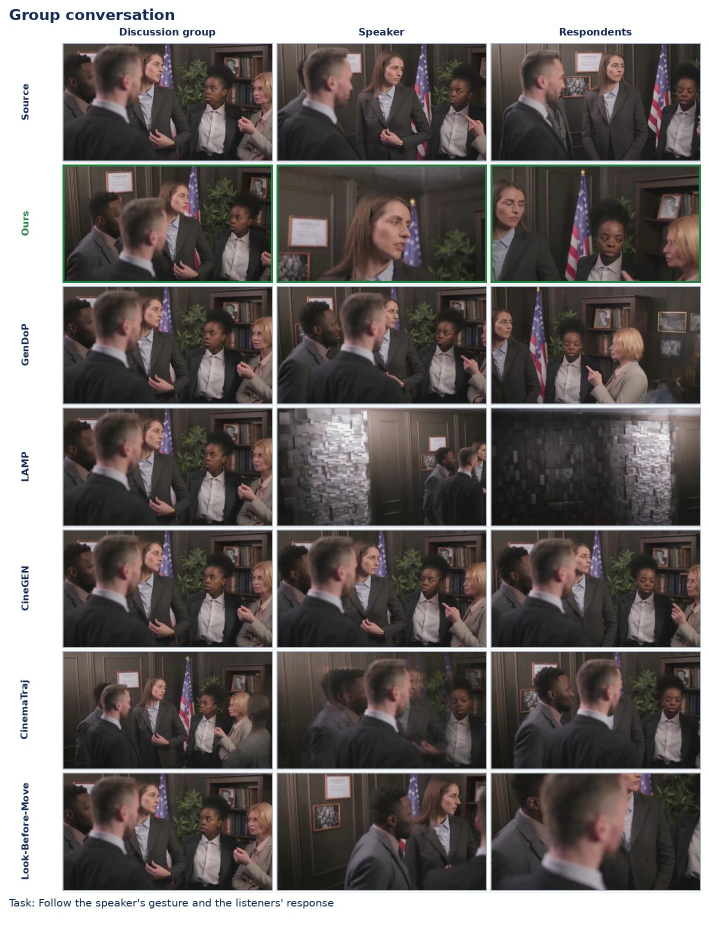}
\vspace{-4pt}
\caption{\textbf{More qualitative comparisons, part 2/10.} We compare our agent with GenDoP, LAMP, CineGEN, CinemaTraj, and Look-Before-Move on a group conversation. Columns follow the discussion group, the central speaker, and the responding participants at matched source times. Rows show the source video and each method's synthesized views; green borders mark ours.}
\label{fig:supp_method_4478885}
\end{figure}

\begin{figure}[!htbp]
\centering
\includegraphics[width=\textwidth,height=0.90\textheight,keepaspectratio]{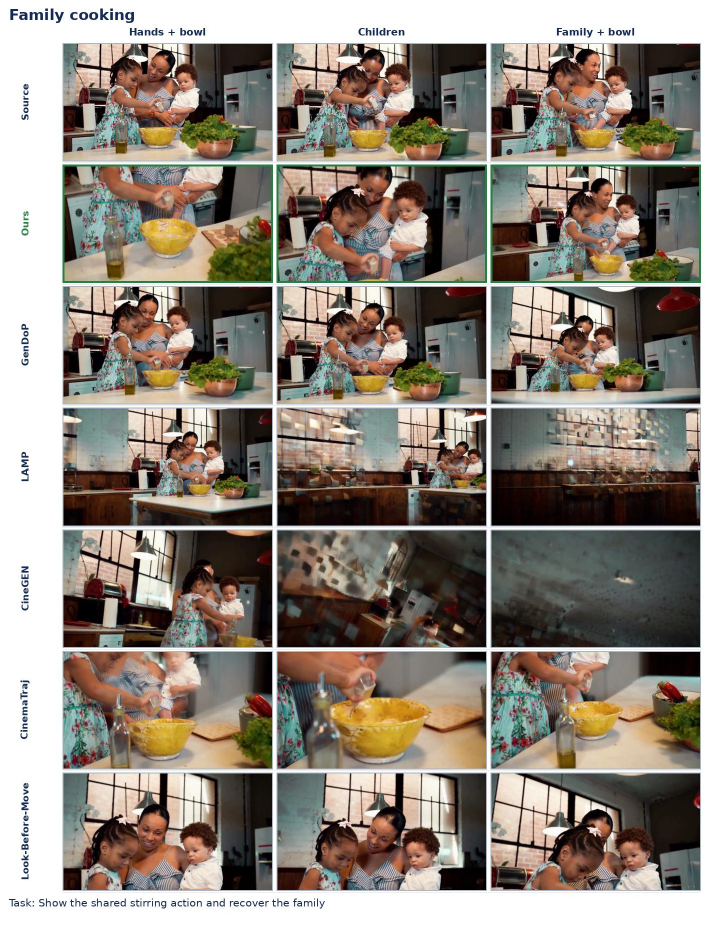}
\vspace{-4pt}
\caption{\textbf{More qualitative comparisons, part 3/10.} We compare our agent with GenDoP, LAMP, CineGEN, CinemaTraj, and Look-Before-Move on family cooking. Columns show matched moments emphasizing the stirring hands and bowl, the children at the bowl, and the family arrangement. Rows show the source video and each method's synthesized views; green borders mark ours.}
\label{fig:supp_method_5593000}
\end{figure}

\begin{figure}[!htbp]
\centering
\includegraphics[width=\textwidth,height=0.90\textheight,keepaspectratio]{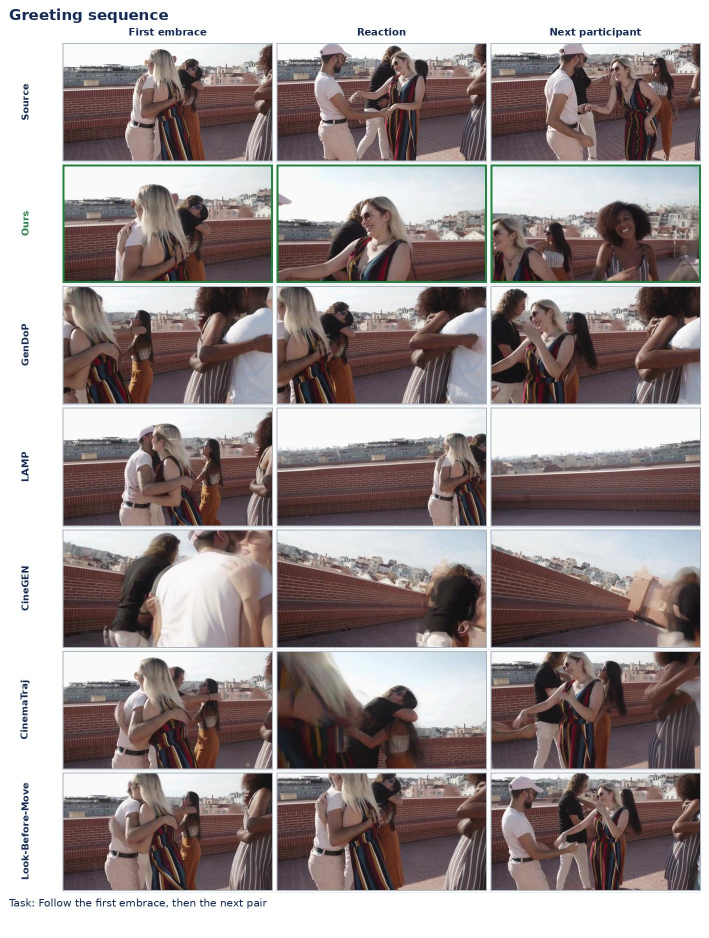}
\vspace{-4pt}
\caption{\textbf{More qualitative comparisons, part 4/10.} We compare our agent with GenDoP, LAMP, CineGEN, CinemaTraj, and Look-Before-Move on a greeting sequence. Columns show the first embrace, the blonde participant's reaction, and the shift toward the next participant at matched source times. Rows show the source video and each method's synthesized views; green borders mark ours.}
\label{fig:supp_method_5935397}
\end{figure}

\begin{figure}[!htbp]
\centering
\includegraphics[width=\textwidth,height=0.90\textheight,keepaspectratio]{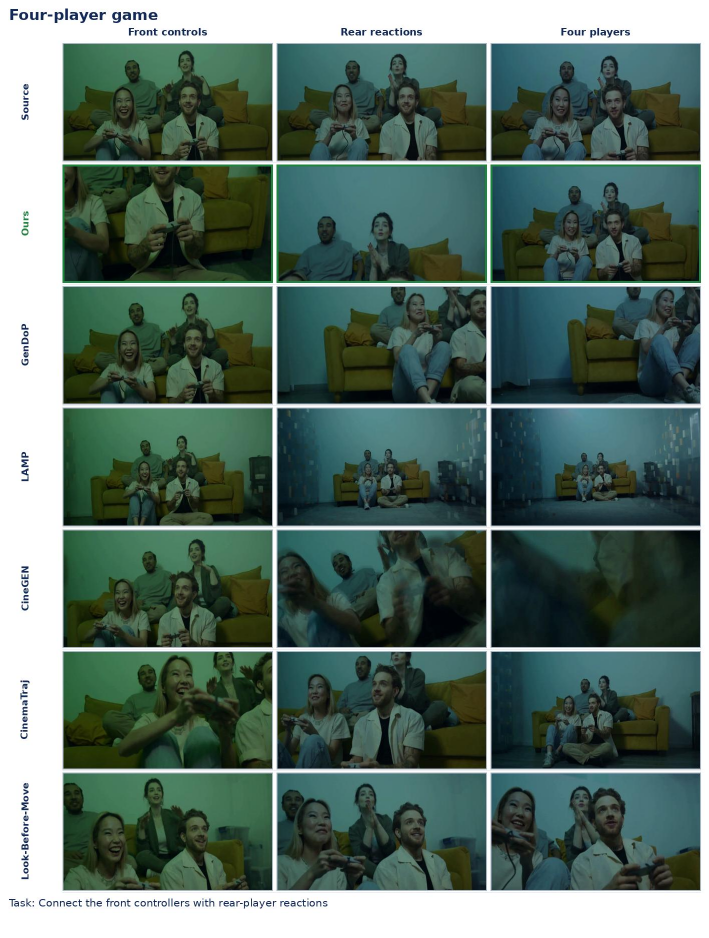}
\vspace{-4pt}
\caption{\textbf{More qualitative comparisons, part 5/10.} We compare our agent with GenDoP, LAMP, CineGEN, CinemaTraj, and Look-Before-Move on a four-player game. Columns show the front controllers, the rear players' reactions, and the four-player composition at matched source times. Rows show the source video and each method's synthesized views; green borders mark ours.}
\label{fig:supp_method_7856709}
\end{figure}

\begin{figure}[!htbp]
\centering
\includegraphics[width=\textwidth,height=0.90\textheight,keepaspectratio]{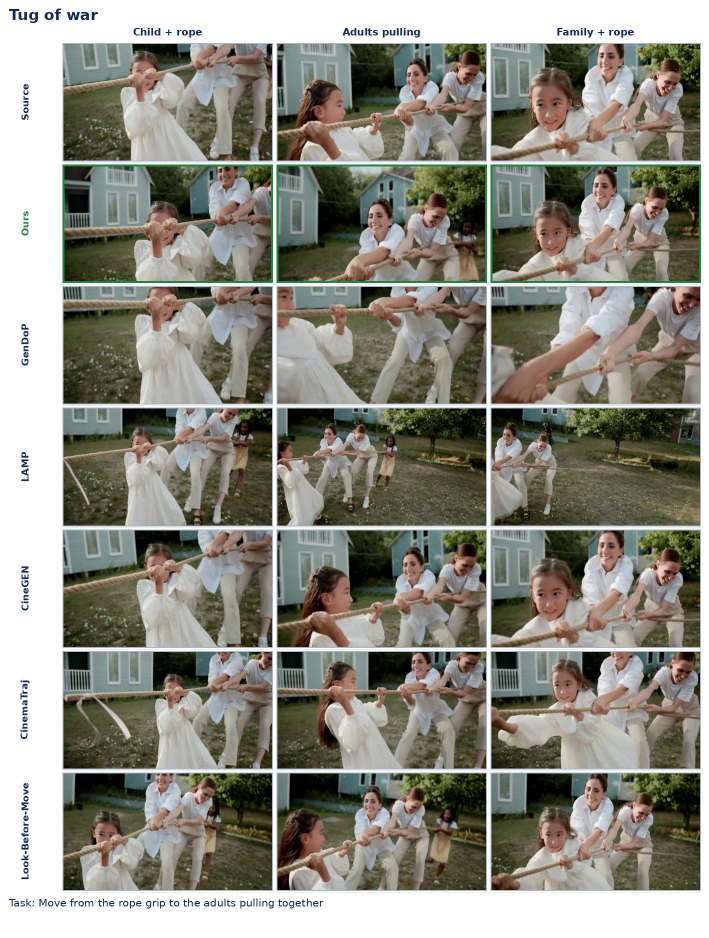}
\vspace{-4pt}
\caption{\textbf{More qualitative comparisons, part 6/10.} We compare our agent with GenDoP, LAMP, CineGEN, CinemaTraj, and Look-Before-Move on tug of war. Columns align matched source times as emphasis shifts from the child's rope grip to the adults pulling and then the family along the rope. Rows show the source video and each method's synthesized views; green borders mark ours.}
\label{fig:supp_method_10037763}
\end{figure}

\begin{figure}[!htbp]
\centering
\includegraphics[width=\textwidth,height=0.90\textheight,keepaspectratio]{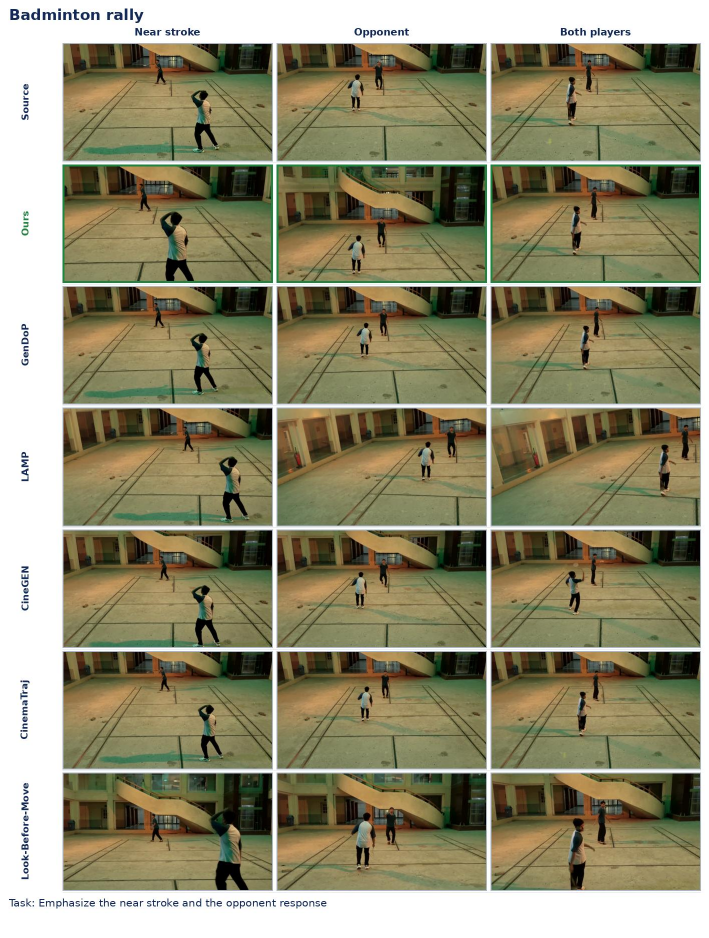}
\vspace{-4pt}
\caption{\textbf{More qualitative comparisons, part 7/10.} We compare our agent with GenDoP, LAMP, CineGEN, CinemaTraj, and Look-Before-Move on a badminton rally. Columns align the near player's stroke, the far player's response, and the two-player view at matched source times. Rows show the source video and each method's synthesized views; green borders mark ours.}
\label{fig:supp_method_35087073}
\end{figure}

\begin{figure}[!htbp]
\centering
\includegraphics[width=\textwidth,height=0.90\textheight,keepaspectratio]{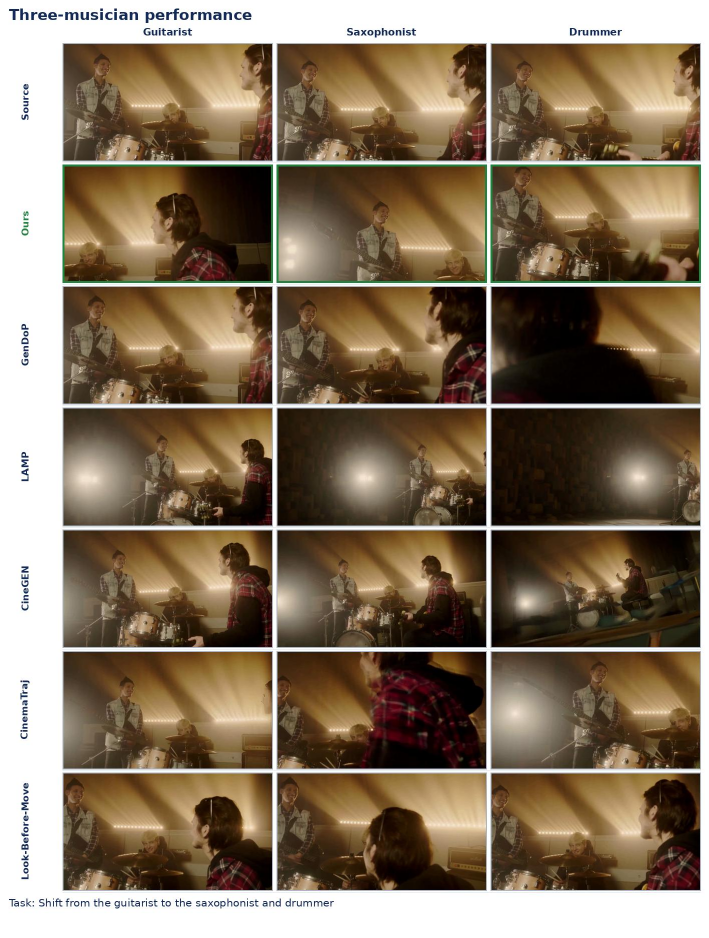}
\vspace{-4pt}
\caption{\textbf{More qualitative comparisons, part 8/10.} We compare our agent with GenDoP, LAMP, CineGEN, CinemaTraj, and Look-Before-Move on a three-musician performance. Columns align matched source times as attention shifts from the guitarist to the saxophonist and drummer. Rows show the source video and each method's synthesized views; green borders mark ours.}
\label{fig:supp_method_6274527}
\end{figure}

\begin{figure}[!htbp]
\centering
\includegraphics[width=\textwidth,height=0.90\textheight,keepaspectratio]{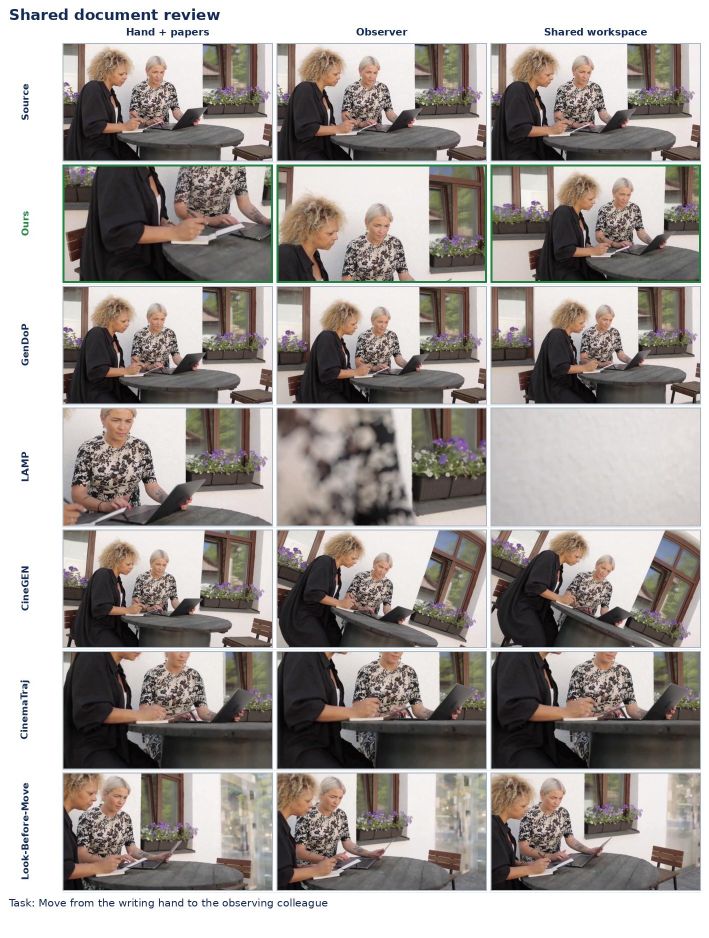}
\vspace{-4pt}
\caption{\textbf{More qualitative comparisons, part 9/10.} We compare our agent with GenDoP, LAMP, CineGEN, CinemaTraj, and Look-Before-Move on shared document review. Columns show matched moments for the writing hand and papers, the observing colleague, and the shared workspace. Rows show the source video and each method's synthesized views; green borders mark ours.}
\label{fig:supp_method_8485159}
\end{figure}

\begin{figure}[!htbp]
\centering
\includegraphics[width=\textwidth,height=0.90\textheight,keepaspectratio]{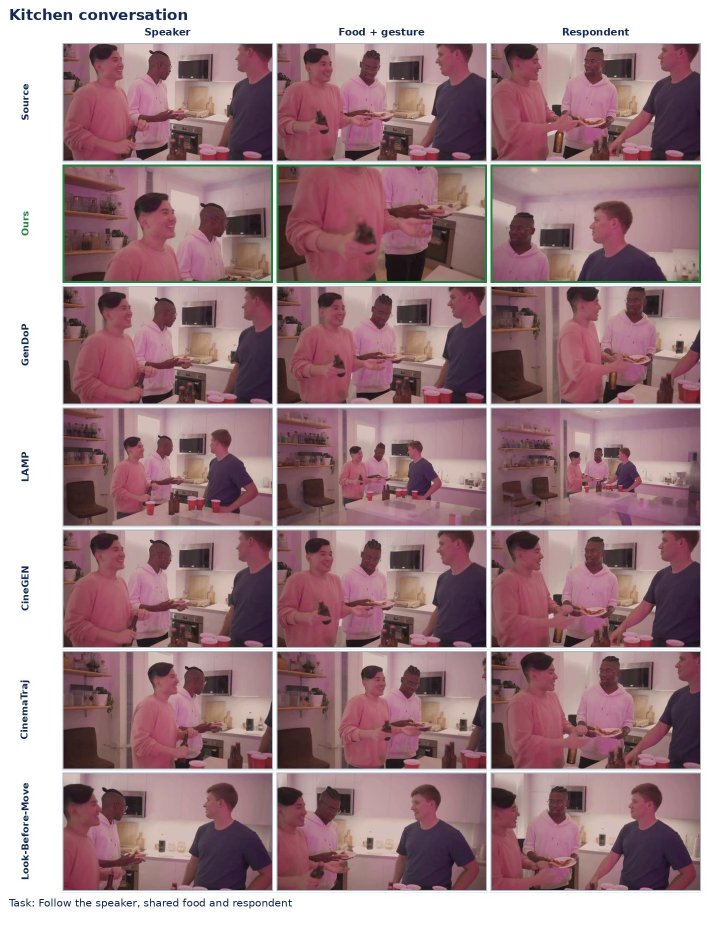}
\vspace{-4pt}
\caption{\textbf{More qualitative comparisons, part 10/10.} We compare our agent with GenDoP, LAMP, CineGEN, CinemaTraj, and Look-Before-Move on a kitchen conversation. Columns follow the speaker, the shared food and gesture, and the respondent at matched source times. Rows show the source video and each method's synthesized views; green borders mark ours.}
\label{fig:supp_method_8757809}
\end{figure}

Figures~\ref{fig:supp_method_4384888}--\ref{fig:supp_method_8757809} compare matched moments across methods at matched source times for the three task stages. Baseline trajectories frequently drift into empty background during large moves (e.g., LAMP in Figs.~\ref{fig:supp_method_4478885}, \ref{fig:supp_method_5593000}, and~\ref{fig:supp_method_8485159}), miss local hand--object actions (Figs.~\ref{fig:supp_method_4384888} and~\ref{fig:supp_method_5593000}), or lose participants when shifting attention, whereas our agent follows each requested action and recovers the group. Green borders mark ours.

\clearpage
\subsection{Grounding and Review Ablations}
\begin{figure}[!htbp]
\centering
\includegraphics[width=\textwidth,height=0.75\textheight,keepaspectratio]{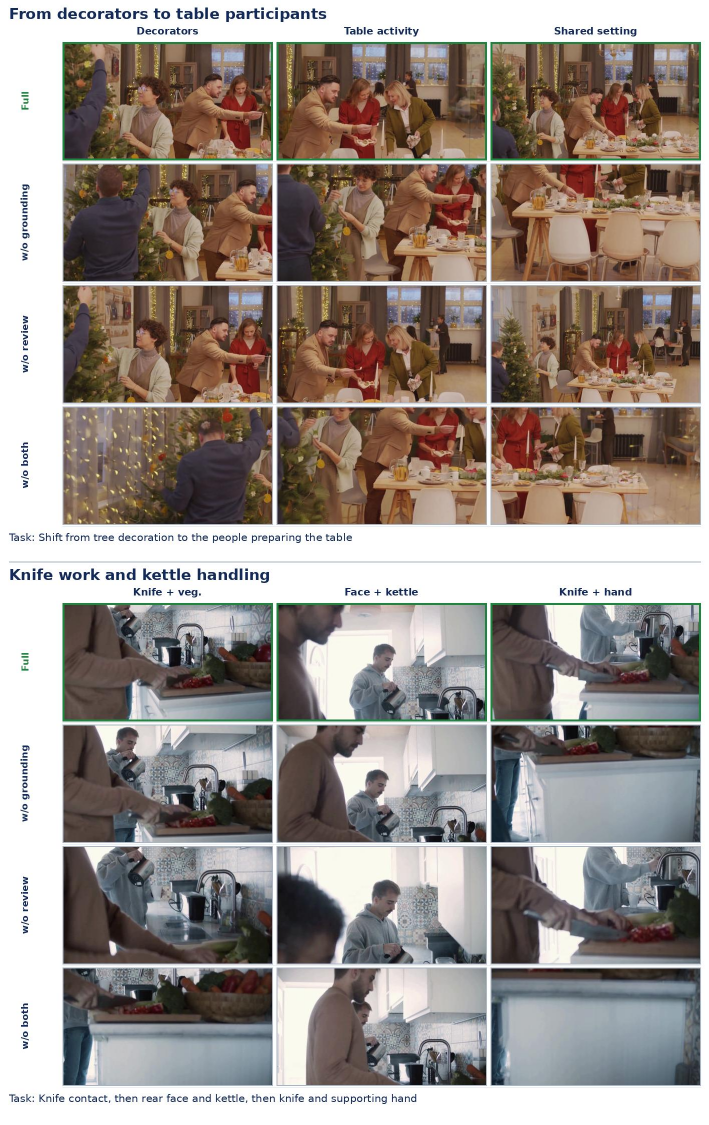}
\vspace{-4pt}
\caption{\textbf{More qualitative ablation results, part 1/3.} We compare the full agent (Full) with variants without dynamic grounding (w/o grounding), without review and correction (w/o review), and without both (w/o both). The upper example shifts attention from tree decorators to table participants and their shared setting. The lower example moves from knife contact with vegetables to the rear worker's face and kettle, then back to cutting. Rows show the four variants and columns align task stages at matched source times; green borders mark Full.}
\label{fig:supp_ablation_a}
\end{figure}

\begin{figure}[!htbp]
\centering
\includegraphics[width=\textwidth,height=0.90\textheight,keepaspectratio]{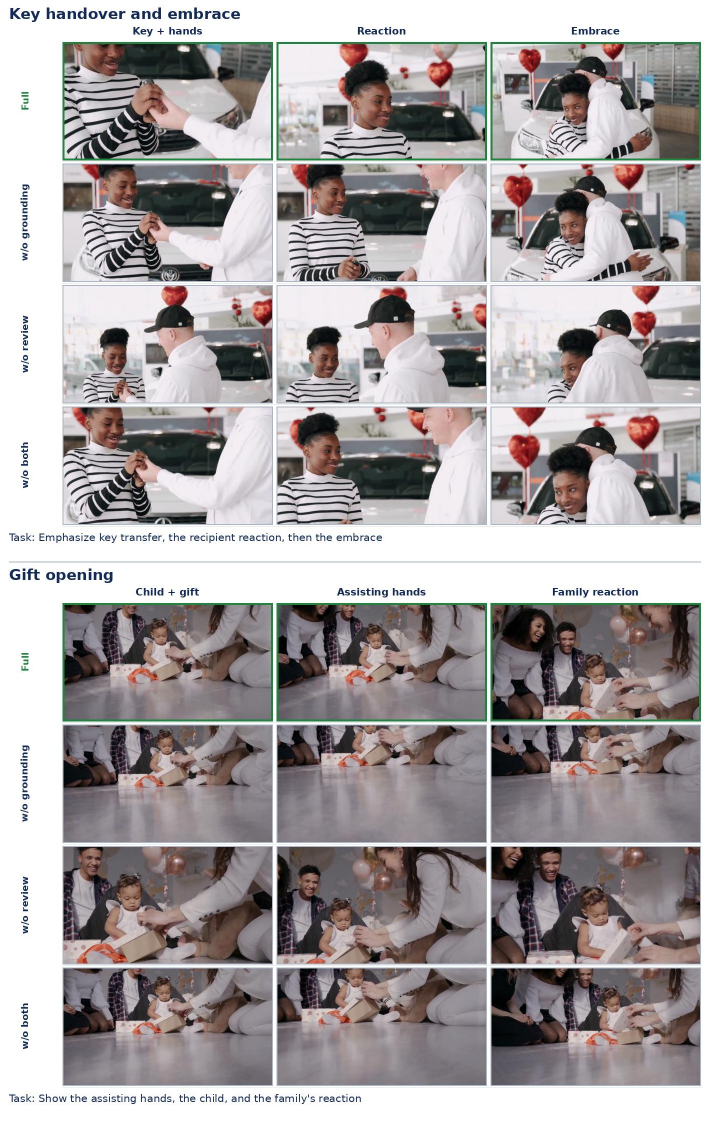}
\vspace{-4pt}
\caption{\textbf{More qualitative ablation results, part 2/3.} We compare Full, w/o grounding, w/o review, and w/o both to examine dynamic grounding and review-based correction. The upper example follows a key handover, the recipient's expression, and the embrace. The lower example follows gift opening, assisting hands, and family reactions. Rows show the four variants and columns align task stages at matched source times; green borders mark Full.}
\label{fig:supp_ablation_b}
\end{figure}

\begin{figure}[!htbp]
\centering
\includegraphics[width=\textwidth,height=0.90\textheight,keepaspectratio]{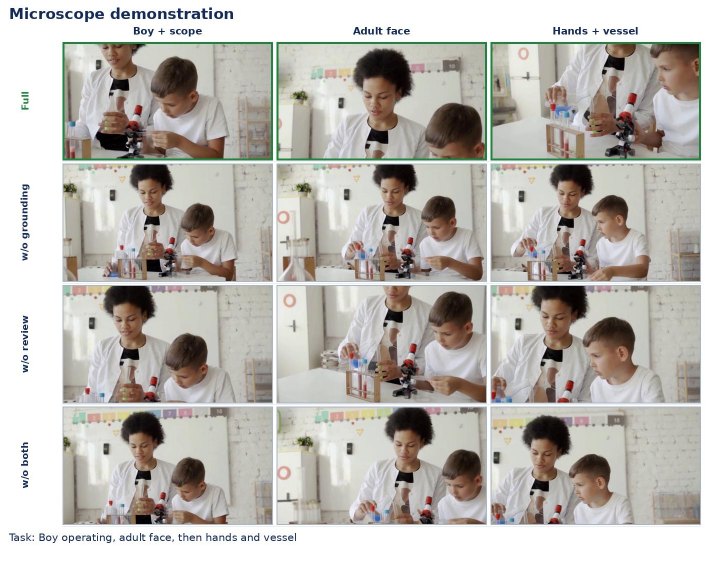}
\vspace{-4pt}
\caption{\textbf{More qualitative ablation results, part 3/3.} We compare Full, w/o grounding, w/o review, and w/o both to examine dynamic grounding and review-based correction in a microscope demonstration. Columns shift attention from the boy using the microscope to the adult's face and then the hands and vessel. Rows show the four variants and columns align task stages at matched source times; green borders mark Full.}
\label{fig:supp_ablation_c}
\end{figure}

Figures~\ref{fig:supp_ablation_a}--\ref{fig:supp_ablation_c} compare Full with variants without dynamic grounding (w/o grounding), without review and correction (w/o review), and without both (w/o both) at matched source times; green borders mark Full. Removing dynamic grounding misses fine action regions (e.g., knife contact or assisting hands), while removing review leaves framing offsets uncorrected (e.g., the review-driven gaze correction in the lower example of Fig.~\ref{fig:supp_ablation_a}).

\end{document}